# From Rule-based to Agentic: LLM Agent based Real-time Control of Urban Drainage Systems for Flooding and Sewer Overflow Mitigation

Wenchong Tian[a,c], Shixun Li[a,c], Ruozhou Lin[a,c], Peng Ye[d], Zhiyu Zhang[a,c], Shi Tao[a], Yu-Qi Wang[a,c], Ling Shi[e], Mingfu Guan[f], Haoran Duan[f]*, Zhiguo Yuan[a,b]*

[a] *School of Energy and Environment, City University of Hong Kong, Hong Kong SAR, China.*

[b] *State Key Laboratory of Marine Pollution, City University of Hong Kong, Hong Kong SAR, China.*

[c] *Shenzhen Research Institute, City University of Hong Kong, China.*

[d] *The Chinese University of Hong Kong, Hong Kong SAR, China.*

[e] *Department of Electronic and Computer Engineering, The Hong Kong University of Science and Technology, Hong Kong SAR, China.*

[f] *Department of Civil Engineering, The University of Hong Kong, Hong Kong SAR, China.*

* Corresponding authors

E-mail address:

Prof. Zhiguo Yuan (zhigyuan@cityu.edu.hk);

Prof. Haoran Duan (haoran.duan@hku.hk)

**Abstract**

Real-time control (RTC) for urban drainage systems (UDS) remains hampered by sparse in-network sensing. While heterogeneous multi-modal meteorological early warnings often contain decision-critical information for UDS RTC, none of the existing approaches can exploit such externally-sourced information within the control workflow. Here we introduce AgenticAI-Control, an agentic artificial intelligence framework able to assimilate multi-source meteorological early warnings in RTC. Two coupled large language model agents are integrated to generate control strategies by retrieving meteorological early-warnings and fusing them with domain engineering expertise. Validated via simulation studies of a combined sewer catchment in eastern China across 10 real-world rainfall events, the AgenticAI-Control achieved 69.1-90.6% reduction in aggregated flood and overflow volumes compared to the rule-based baseline control. Rigorously standardized prompt engineering and modular workflow constraints substantially compress performance disparities among four distinct LLMs. Our work establishes AgenticAI as a viable tool to unify multi-source meteorological monitoring and practitioner domain knowledge, surmounting longstanding monitoring limitations of UDS RTC and delivering cost-efficient, robust intelligent control solutions for sensor-sparse sewer infrastructures.

## Main

Real-time control (RTC) of urban drainage systems (UDS) dynamically modulates hydraulic infrastructure assets (e.g. gates, pumps, and retention basins) based on real-time hydrological and hydraulic measurements for the mitigation of sewer overflows and urban flooding [1, 2, 3]. It constitutes an indispensable tool for the safe and optimal operation of UDS, particularly under the pressure of compounding pressures of climate change and ongoing urbanization [4, 5, 6, 7].

RTC for UDS has historically relied on simple rule-based control. In recent decades, advanced RTC methodologies, such as model predictive control (MPC) [5, 8, 9] and deep reinforcement learning (DRL) [10, 11, 12, 13] have emerged as leading approaches that exhibit desirable properties and superior control performance [14]. This evolutionary trajectory has witnessed a sharp rise in control-algorithm complexity, progressive improvements in control performance, an expansion of controlled assets from standalone facilities to coupled multi-system configurations, and continuous shifts in control objectives. All these achievements attest to the advancing maturity and development of real-time optimal control in UDS.

Such control strategies are critically reliant on comprehensive monitoring data from the underlying drainage network and massive computational resources for running control algorithms. For instance, practical deployments of standard MPC require complete real-time system states as initial conditions for system dynamic prediction, as well as extensive optimization computations to identify the optimal control action [8, 3]. Likewise, DRL achieves optimal control outcomes only when supported by sufficient high-quality monitoring data and computationally expensive training of control models [15, 16]. However, existing UDS monitoring frameworks are restricted to limited sensors of hydrological, hydraulic, and water quality features [17, 18, 19]. The paucity of monitoring information becomes one of the most pressing challenges confronting RTC in UDS [20]. Expanded sensor deployment represents the most straightforward solution, yet budgetary and operational constraints render large-scale sensor networks across urban drainage systems largely unsustainable. Other types of readily accessible information, such as meteorological early warnings, on-site operator feedback, and operator experience, are highly valuable for UDS RTC decision-making. However, these data feature heterogeneous multi-source modalities, including textual narratives, categorical rating data, and even satellite imagery, which are inaccessible to existing RTC systems.

Here we develop a novel intelligent RTC framework for UDS, complementary to conventional RTC frameworks. Leveraging agentic artificial intelligence (AgenticAI), it combines large language models (LLMs) with data, functional tools, workflow orchestration and safety guardrails to exploit early warning information and human expertise. Supported by its core competencies in multi-source data processing and automated tool invocation (e.g. web browsing), the system acquires and parses diverse multi-source information in real time, including regional meteorological early warnings and sewer network status. Therefore, it directly addresses the core limitations of homogeneous data modalities at the data acquisition stage. Furthermore, embedded with a comprehensive knowledge base covering site-specific regulations and operator experience, the system systematically converts multi-source information into executable decision-making insights while tightly coupling external perception data with the RTC framework to build an integrated closed-loop control workflow, address existing data assimilation bottlenecks. Combining all these processes, this system reduce overflows and flooding, thereby demonstrating the feasibility of a warning-control framework centred on multi-source and multimodal data. It also offers a viable pathway to overcoming the monitoring and data assimilation bottleneck that has constrained the deployment of RTC in UDS.

**AgenticAI-enabled RTC framework for UDS RTC**

AgenticAI is a large language model (LLM) combined with an agent harness, which is the surrounding infrastructure that gives the LLM data, tools, orchestration, and safety. This endows LLMs with a comprehensive repertoire of capabilities and is able to automatically retrieve and parse web-hosted content, analyze multi-source data streams, and synthesize multi-source heterogeneous information for structured reasoning and interpretation. Such paradigms have already been maturely deployed across a broad range of domains, including robot control and autonomous driving [21, 22, 23, 24], bio/chemical and medicine research [25, 26, 27, 28], as well as information retrieval and tool invocation in environmental systems [29, 30, 31, 32]. We for the first time propose an AgenticAI-enabled RTC framework for UDS, named AgenticAI-Control, to augment operational decision-making for UDS.

Two core AgenticAI-empowered modules integrate multi-source monitoring data alongside domain specific engineering expertise into the control workflow (Fig. 1). The first, termed the Multi-source Meteorological Monitor, is responsible for the real-time acquisition and parsing of meteorological early warning information. The second module, designated the Control Guide, performs AgenticAI-supported controls based on parsed contextual information. At each control timestep, the Multi-source Meteorological Monitor is invoked to retrieve real-time meteorological warning. When active rainfall warnings are detected, raw multi-source inputs are standardized into structured forecast outputs and forwarded to the Control Guide, which then synthesizes these contextual inputs with established engineering operational expertise to produce calibrated control directives. A safety guardrails modular is integrated into the Control Guide at the final step. This module adopts a control recommendation caching mechanism to reduce unnecessary LLM invocations when no new monitoring data arrives, thereby mitigating damage to control facilities caused by frequent adjustments of control actions. It also further verifies whether the generated control actions are triggered at appropriate execution timings and conform to relevant specifications in prior knowledge to guarantee safety. If these conditions are not satisfied, a validated benchmark control strategy will be adopted. Though lacking a hard-coded explicit control algorithm, the system functions as a round-the-clock online control expert, responding to newly acquired data informed by operational experience.

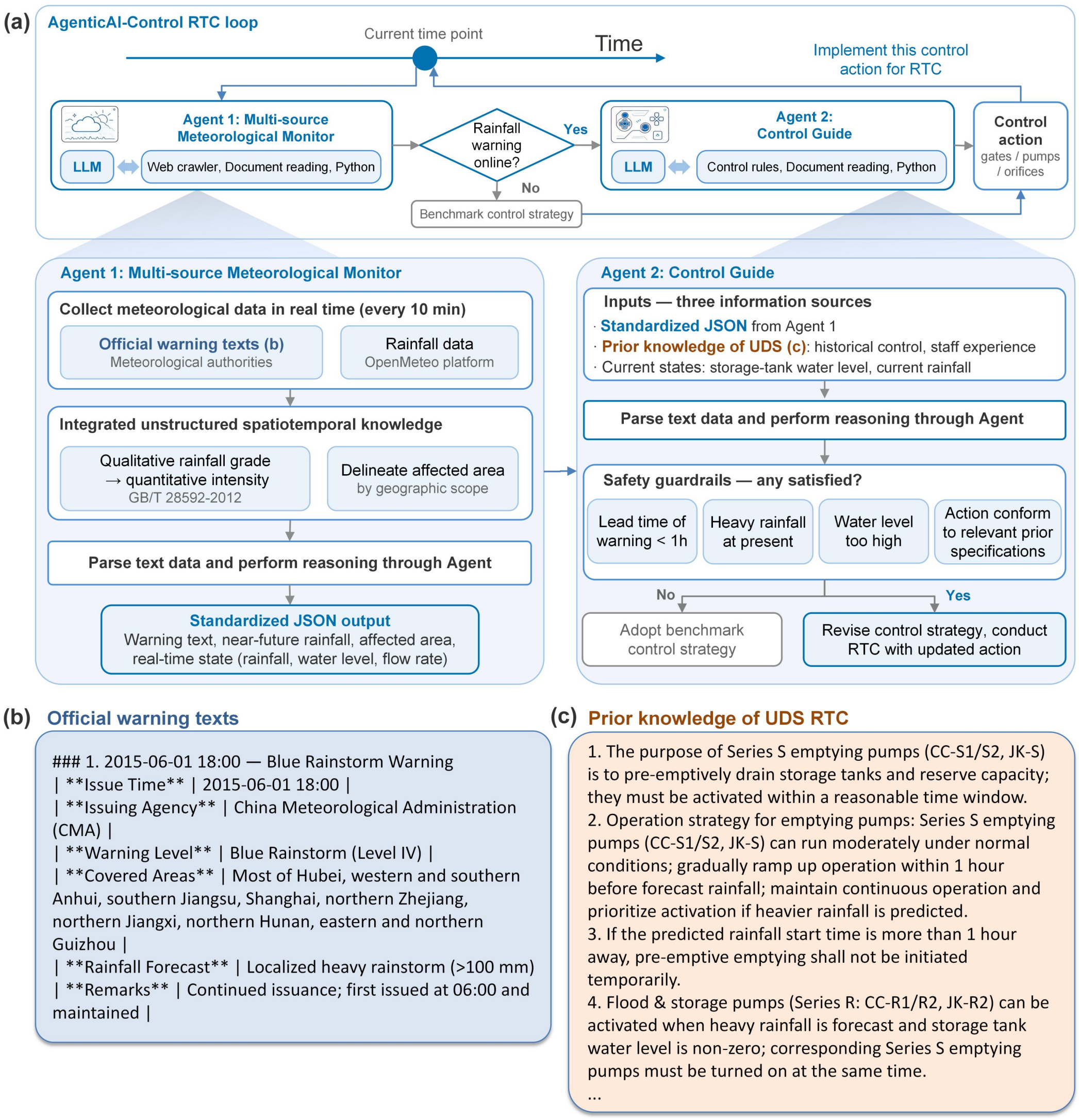

Fig. 1. (a) AgenticAI-Control: a multi-agent RTC framework for UDS via fusion of multi-source monitoring data and engineering expertise. (b) An example of official warning text from China Meteorological Administration (CMA), more data are given in Supplementary Information S6. (c) An example of prior knowledge of UDS RTC, more details are given in the prompt templates of Control Guide (Supplementary Information S2).

## Control performance in flooding and CSO mitigation

The proposed framework is demonstrated and validated via simulation studies of a real-world UDS in a city in eastern China, for which a calibrated Storm Water Management Model (SWMM) is available [6, 33, 34] (Fig. S1 in S4). The system operates as a combined sewer system (CSS), where excessive wet-weather inflows frequently trigger flooding and combined sewer overflows (CSOs). RTC for the two pumping stations within the catchment, housing 7 pumps in total, has been established as a viable strategy to unlock the system's inherent storage capacity and mitigate flooding and CSO. Currently, the pumping stations run on a conventional water-level-triggered rule-based control scheme, designated as the benchmark control (BC) in this study. The BC scheme operates on a set of rule-based logic governed by storage tank water levels. It activates the corresponding pump for drainage once the water level in the storage tank reaches a predefined threshold. The pump is then switched off when the water level drops to the lower set point. The activation water levels for the control rule are provided in the SI (S5).

The study catchment spans a relatively extensive urban area, yet the existing sewer monitoring infrastructure remains sparse with limited spatial coverage. Hourly observed rainfall data for the site, covering the period from 2015 to 2016, were retrieved from the OpenMeteo platform using the study area's geographic coordinates. We cross-referenced these rainfall records against the time windows of official meteorological warnings released via the official portal of the China Meteorological Administration (CMA, https://www.cma.gov.cn/) and China Weather Network (weather.com.cn) (S6). Ten representative rainfall hyetographs and corresponding official warnings were selected as the testbed to evaluate the performance of AgenticAI-Control (Fig. S2 in S4).

For comparison, we developed a model predictive control (MPC) for this case. MPC delivers near-optimal RTC performance. It forecasts forthcoming system states and iteratively tunes control policies using optimization algorithms at every time-step. During each control interval, rainfall forecasts and real-time system states are fed into the SWMM model to simulate system dynamics across the predictive horizon. The optimization solver thereafter searches for a near-optimal control policy that maximizes the control objective over the prediction window, that is, the mitigation of the volume of flooding and CSO. Only the optimized current-interval control action is subsequently executed. The MPC uses a 10 minute control interval, and its prediction horizon for the optimisation algorithm spans three time-steps, corresponding to a total duration of 30 minutes. The Particle Swarm Optimization (PSO) algorithm is employed to solve the optimization problem at each time-step; detailed PSO parameter settings are summarised in S11. To achieve optimal MPC control performance for validation, we assume that the MPC has access to the initial conditions at all locations within the system, and we removed the limit on solution iterations for the MPC optimization problem at each time-step. These simplifying, non-realistic assumptions are introduced to obtain near-optimal control performance for benchmarking against the Agentic- AI control strategy.

The control performance of BC, MPC, and AgenticAI-Control are compared across the 10 rainfall events through detailed volumetric data for flooding and CSO, along with the relative performance improvement ratios (Table 1 and Fig. 2). For all simulations presented in this section, AgenticAI-Control is deployed with DeepSeek-V4-Flash. Performance comparisons across alternative LLM backbones are provided in the subsequent section. AgenticAI-Control outperforms the BC scheme across all rainfall events, with improvement ratios over BC ranging from 69.1% to 90.6%, verifying its efficacy in mitigating wet-weather system impacts. This consistent enhancement demonstrates that the integration of actionable meteorological early warning information and engineering expertise yields pronounced improvements in RTC system performance. Although MPC yields slightly greater improvements (78.9%–91.9%) (Table 1, Fig. 2), its reliance on full network state monitoring renders it impractical in real-world settings. AgenticAI-Control, by comparison, only requires measurements of a limited number of monitoring points across the study area including flow and water level measurements at 13 locations within the sewer network, together with rainfall records for the catchment. Full details are provided in S4. Meanwhile, the runtime latency of AgenticAI-Control is predominantly determined by the millisecond-level LLM API invocation latency (<40 ms), which is substantially lower than the per-timestep optimization overhead of MPC (>1000 s), the latter's overhead is at least 25,000 times that of AgenticAI-Control (Fig. 2(a)).

The control behaviors of BC, MPC, and AgenticAI-Control are fundamentally different (Fig. 2(b), reasoning of different LLMs is provided in S7). BC controls pumps mainly according to variations in forebay storage tanks' water levels, meaning pumps are activated later than the onset of a rainfall event. MPC adjusts the control action according to the prediction results of system state, thus exhibiting a stable and forward-looking control process. AgenticAI-Control executes operational adjustments upon receiving rainfall early warnings and therefore operates pumps in advance prior to rainfall initiation, exhibiting control characteristics intermediate between BC and MPC. Meanwhile, the incorporation of control experience empowers AgenticAI-Control to perform more decisive pump startup operations, achieving a larger drainage volume than BC within the same period. Such lead time and larger drainage volume enable the system to achieve storage before rainfall arrives and reserves sufficient spare capacity. This contributes directly to its improved control performance, thereby further mitigating flooding and overflow volumes.

Table 1. Flooding and CSO volumes ($10^3$ $m^3$) in 10 real rainfall events with BC, MPC and AgenticAI-Control and relative improvement achieved by AgenticAI-Control and MPC (%).

| Event ID | BC ($10^3$ $m^3$) | AgenticAI-Control ($10^3$ $m^3$) | Improvement of AgenticAI-Control to BC (%) | MPC ($10^3$ $m^3$) | Improvement of MPC to BC (%) |
|---|---|---|---|---|---|
| 1 | 34.7 | 8.8 | 74.6% | 2.8 | 91.9% |
| 2 | 65.6 | 6.2 | 90.6% | 3.1 | 95.3% |
| 3 | 161.9 | 30.5 | 81.2% | 23.3 | 85.6% |
| 4 | 124.6 | 25.6 | 79.5% | 19.1 | 84.7% |
| 5 | 25.1 | 4.8 | 80.9% | 3.3 | 86.9% |
| 6 | 113.4 | 26.1 | 76.9% | 23.9 | 78.9% |
| 7 | 146.9 | 23.4 | 84.1% | 22.1 | 85.0% |
| 8 | 179.0 | 45.9 | 74.4% | 23.7 | 86.8% |
| 9 | 207.7 | 64.2 | 69.1% | 34.2 | 83.5% |
| 10 | 107.3 | 28.6 | 73.3% | 16.6 | 84.5% |

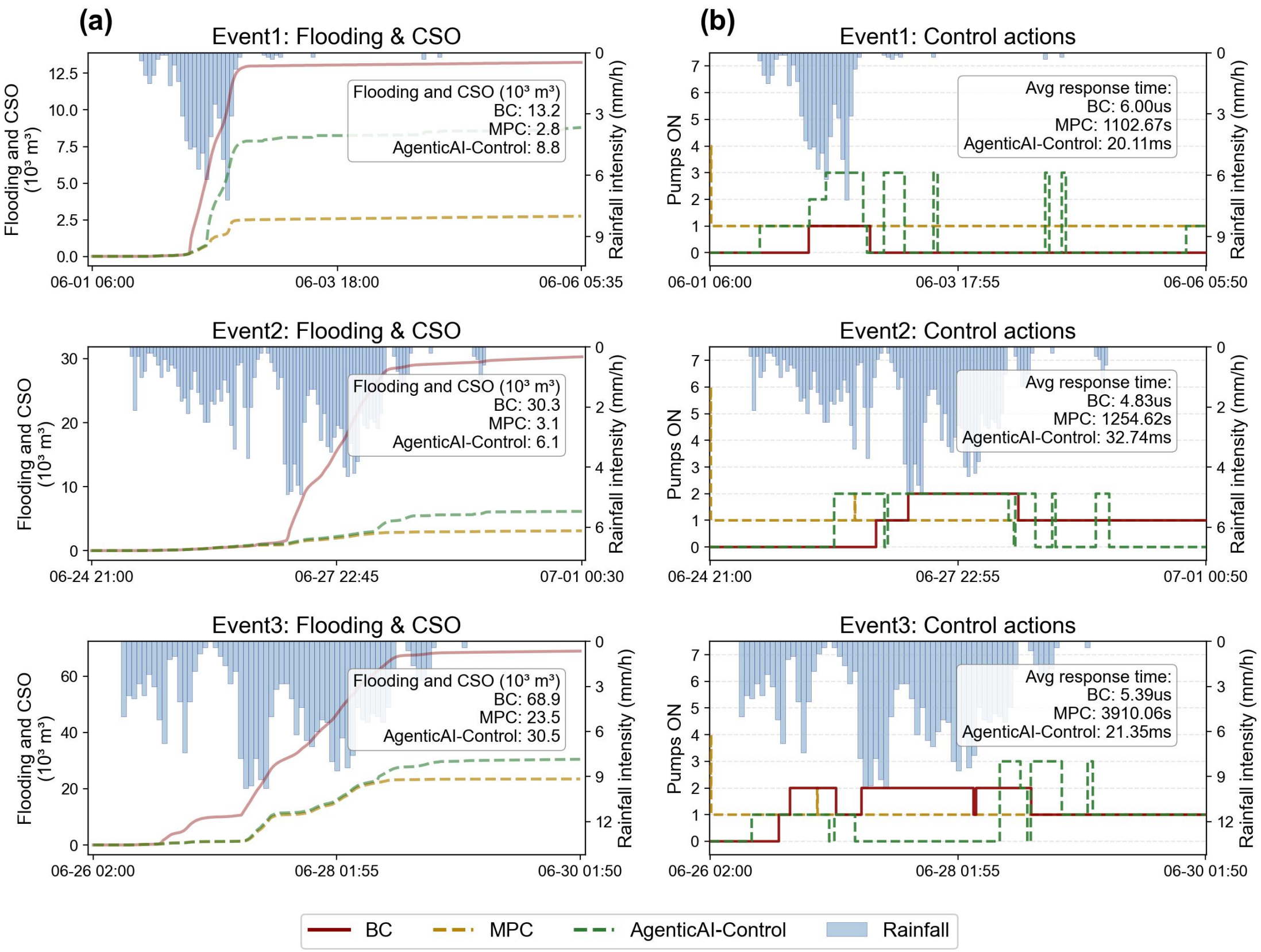

Fig. 2. (a) Control performance of BC and AgenticAI-Control over 3 rainfall events. Colored vertical lines mark the issuance timestamps of graded rainfall early warnings, with distinct hues corresponding to different warning tiers. Warnings in different colors indicate rainfall conditions of varying severity. Blue Warning: The 12-hour accumulated precipitation reaches or exceeds 50 mm, or 50 mm rainfall has occurred with ongoing precipitation. Yellow Warning: The 6-hour accumulated precipitation reaches or exceeds 50 mm, or 50 mm rainfall has occurred with ongoing precipitation. Orange Warning: The 3-hour accumulated precipitation reaches or exceeds 50 mm, or 50 mm rainfall has occurred with ongoing precipitation. Red Warning: The 3-hour accumulated precipitation reaches or exceeds 100 mm, or 100 mm rainfall has occurred with ongoing precipitation. The purple-colored line denotes the issued emergency warning. (b) Control action of BC, MPC, and AgenticAI-Control in Events 1-3. Response times differ among the three control strategies. BC delivers the fastest response as it is rule checking. MPC is the slowest because control actions are obtained from online optimization. AgenticAI-Control acquires control actions via LLM invocation and built-in workflow execution, yielding a performance intermediate between BC and MPC. Fig. 3 shows the detailed API call times. The results in other rainfalls are given in Supplementary Information (S8).

**Comparative performance and uncertainty analysis of various LLMs in AgenticAI-Control**

With the same overall architecture of the AgenticAI-Control, we benchmarked the performance of four distinct LLMs. Since web information retrieval was implemented via dedicated crawler scripts with strictly structured output constraints, the performance variance introduced by different LLMs was primarily evaluated within the parsing and reasoning process of Control Guide module. GLM-4-Flash, Moonshot-v1-8k, and MiniMax-M3 were evaluated along with DeepSeek-V4-Flash used in preceding analyses. OpenAI GPT models and Anthropic Claude models were not included due to network access restrictions. While the four LLMs exhibited varying performance, the overall differences in control efficacy were modest (Fig. 3(a) and S9). The Friedman's non-parametric statistical analysis indicated that LLMs produce no statistically distinguishable differences in end-to-end control performance ($p > 0.05$ in all cases).

All models achieved response latencies below 50 ms and single-step token consumption under 7000 per call, making them fully capable of meeting the real-time requirements of the UDS in this study (Fig. 3(b) and (c)). Specifically, GLM-4-Flash and Moonshot-v1-8k exhibit the lowest token consumption (< 3000 per call) and fastest API call time (< 20 ms), rendering both the most cost-effective options for the task defined in this study. By contrast, MiniMax-M3 delivers comparable control performance but incurs higher token consumption and longer response times, which may be attributed to its default built-in system prompt configuration. Overall, Moonshot-v1-8k is the most suitable choice for this system considering cost efficiency and response speed.

We then conducted uncertainty tests on four types of LLMs to evaluate the influence of hallucination triggered by synonymous expressions. Specifically, multiple rounds of control simulations were carried out by replacing the original prompts with various synonymous expressions (S10). Synonymous substitution of prompts leads to distinguishable variations in the control performance of each AgenticAI-Control scheme, with varying degrees of deviation observed across different rainfall events (Fig. 3(d)). The coefficient of variation (CoV, Eq. 2 in Methodology) under all perturbation conditions is used to evaluate the robustness of each AgenticAI-Control. Moonshot-v1-8k achieved the lowest CoV of 4.8%, comparatively less susceptible to prompt disturbances.

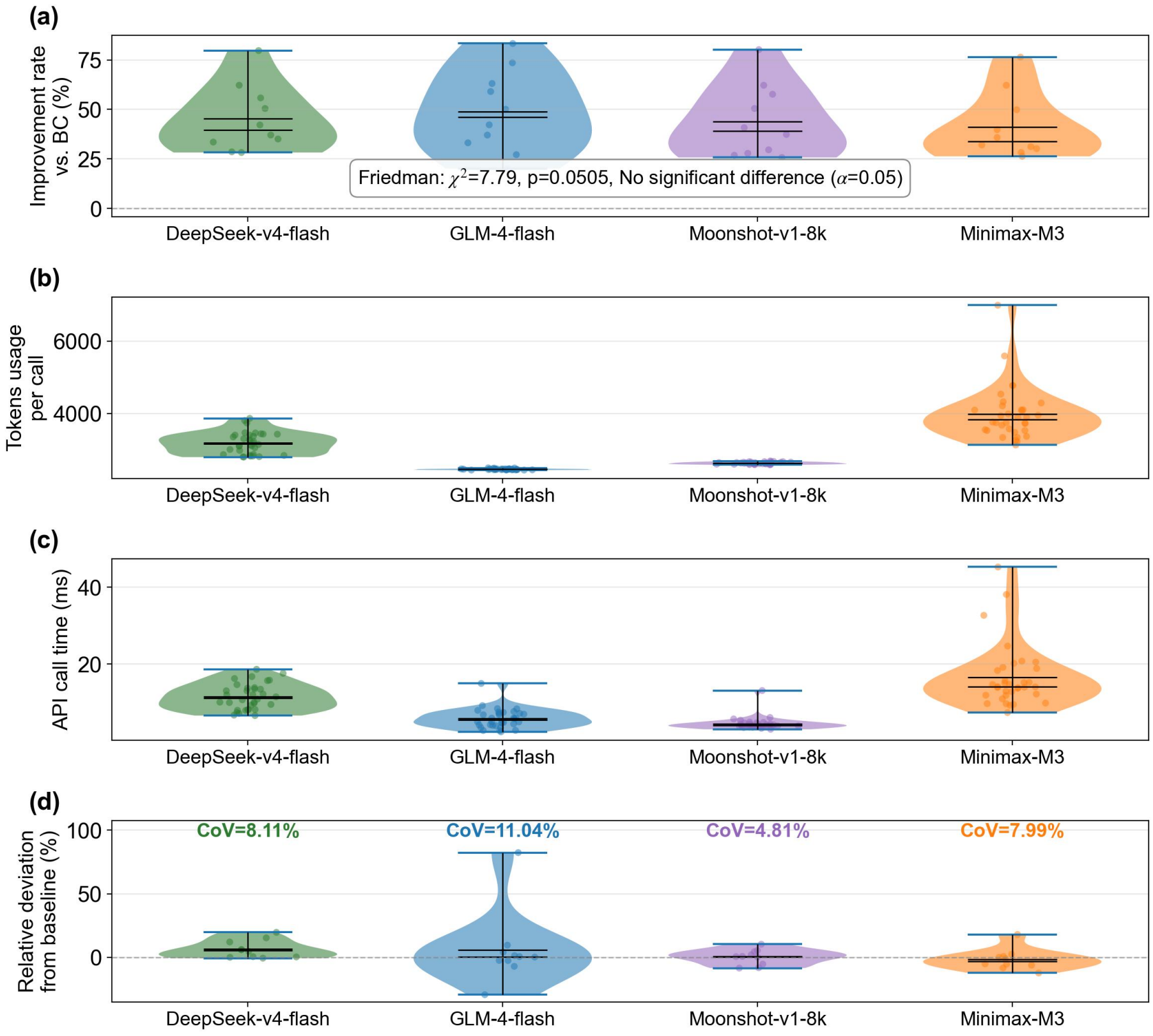

Fig. 3. Improvement rate (a), per-call token usage (b), and response time (c) of AgenticAI-Control with respect to each LLM over 10 rainfall events. (d) Uncertainty analysis and the coefficient of variation (CoV) of AgenticAI-Control under synonym variants over 10 rainfall events.

## Discussion

Over the past five decades, control strategies for drainage systems have evolved substantially: from passive response to proactive real-time operation, and from fixed-rule expert systems to dynamically optimized online control. Nevertheless, all prior RTC implementations in drainage networks have been universally constrained by limited monitoring coverage and insufficient computational capacity. These bottlenecks have prevented RTC from delivering its anticipated performance and hindered its widespread field deployment. The advent of large language models and AI agents offers a transformative opportunity. Leveraging their multimodal reasoning capability, we can seamlessly integrate heterogeneous multi-source real-time observations and rich operational experience into closed-loop control frameworks. This integration enriches the information base for real-time decision-making and reduces reliance on exhaustive online computation, thereby overcoming long-standing limitations of conventional control paradigms and opening a new chapter for UDS RTC.

Herein, we present AgenticAI-Control, an agentic artificial intelligence framework devised to tackle the bottleneck of multi-source meteorological data assimilation that restricts on-site implementation of RTC for UDS. The framework adopts a dual-coupled LLM agent architecture. It autonomously crawls and semantically interprets heterogeneous multi-source meteorological early warnings released by official meteorological platforms, then fuses such predictive information with field operational experience to generate pre-emptive pump regulation strategies. From a broader perspective, this research verifies AgenticAI as an innovative paradigm to integrate scattered multi-source meteorological monitoring data with engineering practitioners' expertise, fundamentally resolving the long-standing insufficient monitoring issue that hinders the popularization of UDS RTC. Distinct from traditional algorithms that merely pursue performance optimization, this framework leverages free publicly accessible meteorological warning data rather than expensive large-scale sensor renovation, thus providing an economical intelligent control scheme applicable to sewer facilities with sparse sensors worldwide.

When benchmarked against the best-performing MPC, AgenticAI-Control delivers relatively near-optimal control performance with remarkable low-latency real-time response and less online hydraulic monitoring, which qualify it as one of the highly competitive RTC strategies. The core driver of the superior performance of AgenticAI-Control lies in the effective utilization of high-quality rainfall early warning information and control engineering experience. By delegating meteorological analysis to specialized authorities, highly accurate and reliable warning information can be obtained at comparatively low cost. When paired with established engineering operational experience, these reasonably accurate forecasts enable control performance comparable to both state-of-the-art RTC approaches. This finding carries important practical implications: optimization of control algorithms alone may not represent a cost-effective strategy for advancing UDS RTC performance. Instead, comparable control outcomes can be achieved through the synergy of accurate early warnings, targeted system monitoring, and operational expertise grounded in a mechanistic understanding of UDS.

The outcomes in Fig. 3 indicate that distinct LLMs yield no statistically distinguishable differences in performance when deployed within the same strict and comparatively fixed agent workflow framework. This suggests that state-of-the-art LLMs are not required for the proposed AgenticAI-Control. Its well-designed prompt and workflow can yield robust control performance. This finding aligns with existing literature on AgenticAI, which establishes that model selection has limited impact on end-to-end system performance when deployed within well-architected frameworks with standardized prompts and structured workflow pipelines [35, 36]. Accordingly, robust system architecture design can effectively compensate for inherent limitations of individual LLMs in practical deployments, which also provides a viable pathway for offline small-size LLM application scenarios. Nevertheless, the variations in prompt design still affect the overall control efficacy, implying that crafting more precise prompts is important for the proposed framework.

## Methods

### *AgenticAI enabled meteorological early warning and UDS RTC*

#### *Multi-source Meteorological Monitor*

The overall architecture of Multi-source Meteorological Monitor comprises two tightly integrated steps enabled by AgenticAI: monitoring acquisition and multi-source monitoring information processing. A detailed schematic of the framework is provided in Fig. 1(b).

The first step retrieves meteorological early warning data from public web portals. Operating at a 10-minute temporal resolution, it conducts rolling monitoring of multi-source public data streams, ingesting official warning texts issued by national and local meteorological authorities and fine-resolution hourly rainfall estimates from the OpenMeteo platform. Upon detection of rainfall-related warning signals, the system automatically extracts warning metadata and identifies three core information categories from unstructured text via LLM-driven natural language semantic parsing. These include warning validity period, affected geographic extent, and rainfall intensity grade. Subsequently, the second step performs structured processing of the acquired information. Qualitative rainfall grades are converted into quantitative rainfall intensity values in compliance with warning level from meteorological authorities, and the spatial boundary of rainfall impact is preliminarily delineated according to the geographic scope defined in the warning. By fusing rainfall warning records with emergency incident point data to pinpoint event locations. In addition to the information mentioned above, real-time system states data obtained through online monitoring are also sent to the AgenticAI. These include water levels and flow rates at selected nodes and pipes within the UDS, as well as contemporaneous rainfall measurements. The final output encompasses fields including basic rainfall warning attributes, estimated rainfall intensity and spatial impact extent, and current monitored states, forming a structured supplementary regional monitoring.

The entire end-to-end workflow runs fully autonomously, enabled by the synergistic interplay of prompt engineering and orchestrated workflow invocation. Retrieval and crawling of public meteorological information are performed by the agent via dedicated web search utilities and prepared python script; standardized rainfall grade alignment and semantic information extraction leverage the LLM's native natural language understanding capabilities to parse national standard document and retrieved text; and the final structured dataset is cooperatively generated through LLM-powered semantic parsing. To support data interoperability within the downstream AgenticAI, all monitoring data are standardized for storage and transmission in JavaScript Object Notation (JSON), a lightweight, text-based, language-agnostic structured data format. Full details of the core prompt templates are provided in the Supplementary Information (S1).

***Control Guide***

This subsequent module deploys a second dedicated AgenticAI instance to deliver a comprehensive control strategy by integrating empirical human operational control expertise, achieving an integration of domain knowledge, supplementary monitoring inputs, and intelligent control architectures. Specifically, this module encompasses two core steps: the formulation of tailored control recommendations, and a decision-gating that evaluates the feasibility of the proposed adjustments. The detailed architecture of this module is illustrated in Fig. 1(c).

For the first step, we embed prior domain constraints from UDS RTC practice to confine the AgenticAI's output to a well-defined decision scope: whether to activate pumps for drainage in advance, and which specific pumps to engage based on spatial analysis and on-site system control protocols. First, the JSON-formatted standard outputs are ingested to perform holistic reasoning of the embedded multi-source information. To enhance the depth and fidelity of reasoning and overcome the limitations inherent in single-source data interpretation, domain-specific auxiliary context encoded in natural language are integrated into the reasoning. This contextual knowledge includes UDS topological specifications, which define core network characteristics (including pipe connectivity, node distribution, and pipe diameter parameters), site-specific UDS documentation, encompassing design codes, empirical operation and maintenance manuals, and peer-reviewed research findings for the target catchment. Upon receiving these inputs, the AgenticAI produces control directives through native LLM reasoning. To safeguard the reliability and interpretability of its output, the generated response is required to comprise three core components: recommended adjustments to control parameters, a detailed rationale underpinning the proposed adjustments, and an assessment of the current UDS operational state informed by monitoring.

Since AgenticAI-generated control advice reflects projections of future system trends rather than precisely timed intervention triggers, its deployment must be calibrated against the real-time operational state of the UDS. Accordingly, the second component determines whether adopting the AgenticAI's recommendations is opportune at each control timestep. Centered on the AgenticAI's recommendation, this component implements a logical branching that integrates above output to execute a hierarchical sequential decision-making, enabling precise adaptation of pre-storm drainage strategies. The process operates in two sequential stages. The first-stage screening applies a temporal threshold of one hour prior to forecast rainfall onset. That is, if the current timestep remains more than one hour ahead of the warning window, no strategy adjustment is triggered, and the baseline control continues. Once the system enters the one-hour pre-warning response window, a second-stage state validation is conducted, jointly evaluating real-time rainfall intensity and water levels at storage nodes. When real-time rainfall shows an increasing trend, or key node water levels are already elevated and the system has entered the formal warning window, the AgenticAI recommendations are formally adopted. Conversely, the recommendation is deferred and the baseline control is preserved unchanged.

The entire workflow is fully automated via prompt engineering combining both JSON dataset and auxiliary contextual information. Detailed specifications of the core prompt templates are provided in the Supplementary Information (S2). To prevent redundant LLM invocations by the Control Guide in the absence of incremental monitoring inputs, we embed a control recommendation caching mechanism within the decision-gating pipeline. Specifically, upon the initial acquisition and reasoning of meteorological early warning information, the full set of control recommendations generated at that timestep is persisted in the system cache. As long as no updated meteorological warning data are received and valid cached recommendations remain available, the system refrains from re-invoking the LLM to generate new control outputs. This mechanism eliminates redundant processing of identical information under static warning conditions, thereby reducing unnecessary token consumption.

### *Performance index and uncertainty analysis*

The primary metrics for evaluating control efficacy are reductions in total CSO volume and flooding magnitude. We further define a control performance improvement ratio, calculated as Eq.1, where $flooding_{BC}$ and $CSO_{BC}$ denote the total flooding magnitude and CSO volume yielded by the baseline schemes BC. Correspondingly, $flooding_{AIC}$ and $CSO_{AIC}$ represent the equivalent metrics produced by AgenticAI-Control. A higher percentage value indicates superior control performance.

$$Improvement_rate = \frac{flooding_{AIC} + CSO_{AIC} - (flooding_{BC} + CSO_{BC})}{flooding_{BC} + CSO_{BC}} \times 100\% \quad (1)$$

We conduct an uncertainty analysis of the AgenticAI-Control to evaluate the potential impact of near-synonymous lexical substitutions in prompts on its control efficacy. Specifically, selected terms within the original baseline prompt templates are randomly replaced with their semantically close synonyms to generate distinct prompt variants, which are then implemented in system control simulations. The substitutions span four core descriptive domains: rainfall characteristics, warning issuance mode, affected geographic extent, and system operational state. The full substitution term list is provided in the Supplementary Information (S10). We analyze the control performance of AgenticAI-Control with various LLMs under the interference of perturbed prompts. The coefficient of variation is adopted to quantitatively characterize the fluctuation magnitude of control performance, so as to evaluate the robustness of each AgenticAI-Control. The formula for calculating coefficient of variation is defined in Eq.2, where $\sigma$ is the standard deviation of control performance across diverse perturbation scenarios, and $\mu$ represents the average system output obtained under these perturbed conditions. A lower coefficient of variation corresponds to superior robustness of the overall system.

$$Coefficient_Variation = \frac{\sigma}{\mu} \times 100\% \quad (2)$$

### *Near-optimal control using MPC*

MPC represents a state-of-the-art approach with favourable control performance for RTC research on UDS, and is incorporated as benchmark schemes for comparative evaluation. It achieves near-optimal RTC performance by forecasting future system states and iteratively optimizing control strategies at each timestep via embedded optimization algorithms.

The implemented MPC in UDS relies on three prerequisites: (i) rainfall forecasts; (ii) the availability of flow and water-level measurements across the UDS; (iii) computation time for the optimization solver. In this study, a system prediction model and hydraulic monitoring are constructed by using the correct rainfalls prediction and water flow and level simulation as inputs to the SWMM. The combined volume of flooding and CSO is defined as the optimization objective at each timestep. The complete optimization formulation is presented in Eq. 3, where $r_t$ denotes the stage objective function, aggregated from $flooding(t)$ and $CSO(t)$ over the entire prediction horizon (30 minutes in this case). The control actions $\{a_t, a_{t+\Delta t}, \dots, a_{t+pred_horizon}\}$, spanning from the current timestep to the prediction horizon, serve as the optimization variables. The $a_t$ is applied as the executed control action for the current timestep, following the receding horizon principle of MPC. Term $inflow_t$ corresponds to the system inflow over the time interval. Solving Eq. 3 in this study uses the full hydrodynamic states from SWMM (encompassing water levels $wl_t$ at all network nodes and flow rates $wf_t$ in all conduits) and error-free rainfall forecasts ($rain_t$). The PSO algorithm is adopted as the real-time optimizer. Detailed parameter configurations, alongside computing cost are available in the S11. To guarantee its control performance, we did not impose limit on the computation time of the PSO algorithm during the tests.

$$\max_{\{a_t, a_{t+\Delta t}, \dots, a_{t+pred_horizon}\}} \sum_{\tau=t}^{t+pred_horizon} r_t \tag{3}$$

$$r_t = \begin{cases} \dfrac{1}{1+\dfrac{flooding(t)+CSO(t)}{inflow_t}} - 1, & if\ inflow_t \neq 0 \\ 0, & else \end{cases}$$

$$flooding(t), CSO(t) = SWMM(rain_t, wl_t, wf_t)$$

**Supplementary Information**

Sections S1-S11 are provided in the Supplementary Information.

**Acknowledgements**

This study is supported by RGC/UGC GRF (grant number 11210825) and the National Natural Science Foundation of China (NSFC) (grant number 52400113). Zhiguo Yuan is a Global STEM Professor jointly funded by the Innovation, Technology and Industry Bureau ("ITIB") and Education Bureau ("EDB") of the Government of the Hong Kong Special Administrative Region, China and acknowledges financial support from the Hong Kong Jockey Club for the JC STEM Lab of Sustainable Urban Water Management.

The code and data used in this study has been open-sourced on Github.

Github link: https://github.com/DantEzio/LLMAgent_UDSControl

**Data availability**

Rainfall observations and forecasts (from Open-Meteo), meteorological early warnings (from the China Meteorological Administration, China Weather Network), and the calibrated SWMM model used in this study are publicly available on GitHub.

**Code availability**

The AgenticAI-Control code, prompt templates and MPC benchmark are available on GitHub.

**Author contributions**

- Conceived and designed the experiments: Wenchong Tian, Shixun Li, Ruozhou Lin, Peng Ye, Zhiyu Zhang, Tao Shi, Ling Shi, Mingfu Guan, Haoran Duan.
- Performed the experiments: Wenchong Tian, Shixun Li, Haoran Duan.
- Analyzed the data: Wenchong Tian, Shixun Li, Yu-Qi Wang, Zhiguo Yuan.
- Contributed materials/analysis tools: Wenchong Tian, Shixun Li, Ruozhou Lin, Peng Ye, Yu-Qi Wang, Zhiguo Yuan.
- Wrote the paper: Wenchong Tian, Shixun Li, Tao Shi, Yu-Qi Wang, Ling Shi, Mingfu Guan, Haoran Duan, Zhiguo Yuan.

**Competing interests**

The authors declare no competing interests.

## References

1. Schütze, M., Campisano, A., Colas, H., Schilling, W. & Vanrolleghem, P. A. Real time control of urban wastewater systems—where do we stand today? *Journal of Hydrology* **299**, 335–348 (2004).
2. García, L. *et al.* Modeling and real-time control of urban drainage systems: A review. *Advances in Water Resources* **85**, 120–132 (2015).
3. Huang, Z. *et al.* Pollution-based integrated real-time control for urban drainage systems: a multi-agent deep reinforcement learning approach. *npj Clean Water* **8**, 79 (2025).
4. Lund, N. S. V., Falk, A. K. V., Borup, M., Madsen, H. & Mikkelsen, P. S. Model predictive control of urban drainage systems: A review and perspective towards smart real-time water management. *Critical Reviews in Environmental Science and Technology* **48**, 279–339 (2018).
5. Lund, N. S. V., Borup, M., Madsen, H., Mark, O. & Mikkelsen, P. S. CSO Reduction by Integrated Model Predictive Control of Stormwater Inflows: A Simulated Proof of Concept Using Linear Surrogate Models. *Water Resour. Res.* **56**, (2020).
6. Tian, W., Liao, Z., Zhang, Z., Wu, H. & Xin, K. Flooding and Overflow Mitigation Using Deep Reinforcement Learning Based on Koopman Operator of Urban Drainage Systems. *Water Resources Research* **58**, (2022).
7. Pei, S., Fu, G., Hoang, L. & Butler, D. Robustness of neuro-evolution in urban drainage system control under communication failures: comparing centralized and decentralized control schemes. *Water Research* **282**, 123733 (2025).
8. Zhang, Z., Tian, W., Liao, Z. & Yuan, Z. Differentiable neural network-based models enable gradient-based optimization for model predictive control of urban drainage networks. *Water Research* 125188 (2025).
9. Luo, X. *et al.* A decomposition-coordination approach for distributed real-time control of urban drainage systems. *Water Research* **301**, 126042 (2026).
10. Tian, W. *et al.* Flooding mitigation through safe & trustworthy reinforcement learning. *Journal of Hydrology* **620**, 129435 (2023).
11. Negm, A., Ma, X. & Aggidis, G. Deep reinforcement learning challenges and opportunities for urban water systems. *Water Research* **253**, 121145 (2024).
12. He, Z. *et al.* Deep reinforcement learning control as an innovative approach for urban drainage systems: review and prospects. *Water Research* **284**, 123954 (2025).
13. Huang, Z., Wang, Y. & Dong, X. Dimensions of superiority: How deep reinforcement learning excels in urban drainage system real-time control. *Water Research X* **28**, 100313 (2025).
14. Liao, Z., Zhang, Z., Tian, W., Gu, X. & Xie, J. Comparison of Real-time Control Methods for CSO Reduction with Two Evaluation Indices: Computing Load Rate and Double Baseline Normalized Distance. *Water Resour Manage* https://doi.org/10.1007/s11269-022-03221-1 (2022) doi:10.1007/s11269-022-03221-1.
15. Zhang, M., Xu, Z., Wang, Y., Zeng, S. & Dong, X. Evaluation of uncertain signals' impact on deep reinforcement learning-based real-time control strategy of urban drainage systems. *Journal of Environmental Management* **324**, 116448 (2022).
16. Tian, W., Xin, K., Zhang, Z., Liao, Z. & Li, F. State Selection and Cost Estimation for Deep Reinforcement Learning-Based Real-Time Control of Urban Drainage System. *Water* **15**, 1528 (2023).
17. Oh, J. & Bartos, M. Scalable, adaptive and risk-informed design of hydrological sensor networks. *Nature Water* **3**, 1144–1154 (2025).
18. Lin, R., Tian, W., Qiu, R., Hu, L. & Yuan, Z. Low-cost, data-efficient, on-device soft sensors for sewer flow monitoring—learning from adjacent water level sensors. *Water Research X* **29**, 100415 (2025).
19. Lin, R., Qiu, R., Hu, L., Ding, Y. & Yuan, Z. A low-cost soft sensor for sewer

flow monitoring — Learning from water level measurements in manholes. *Water Research* **274**, 123135 (2025).

20. Crowley, G., Tait, S., Panoutsos, G., Speight, V. & Esnaola, I. Information-theoretic sensor placement for large sewer networks. *Water Research* **268**, 122718 (2025).
21. Zitkovich, B. *et al.* RT-2: Vision-Language-Action Models Transfer Web Knowledge to Robotic Control. in *Proceedings of The 7th Conference on Robot Learning* (eds Tan, J., Toussaint, M. & Darvish, K.) vol. 229 2165–2183 (PMLR, 2023).
22. Hwang, J.-J. *et al.* EMMA: End-to-End Multimodal Model for Autonomous Driving. *Transactions on Machine Learning Research* https://openreview.net/forum?id=kH3t5lmOU8 (2025).
23. Sartore, C. *et al.* Towards shared embodied intelligence in humanoid robots through optimization, development and testing of the human-aware ergoCub robot. *Nat Mach Intell* https://doi.org/10.1038/s42256-026-01272-2 (2026) doi:10.1038/s42256-026-01272-2.
24. Li, X. *et al.* What matters in building vision–language–action models for generalist robots. *Nat Mach Intell* **8**, 158–172 (2026).
25. Pan, Y. *et al.* Multi-agent artificial intelligence designs novel catalysts for ultrafast water purification. *Nature Water* **4**, 630–642 (2026).
26. Boiko, D. A., MacKnight, R., Kline, B. & Gomes, G. Autonomous chemical research with large language models. *Nature* **624**, 570–578 (2023).
27. Qiu, J. *et al.* LLM-based agentic systems in medicine and healthcare. *Nat Mach Intell* **6**, 1418–1420 (2024).
28. Swanson, K., Wu, W., Bulaong, N. L., Pak, J. E. & Zou, J. The Virtual Lab of AI agents designs new SARS-CoV-2 nanobodies. *Nature* **646**, 716–723 (2025).
29. Xu, B. *et al.* Evaluating and Advancing Large Language Models for Water Knowledge Tasks in Engineering and Research. *Environ. Sci. Technol. Lett.* **12**, 289–296 (2025).
30. Wang, J., Fu, G. & Savic, D. EPANET-Agentic: A multi-agent system for natural language-controlled simulations of water distribution networks. *Water Research* **293**, 125433 (2026).
31. Zhai, M. *et al.* WaterRAG: A Multiagent Retrieval-Augmented Generation Framework to Support Water Industry Transitions to Net-Zero. *Environ. Sci. Technol.*

**60**, 11529–11541 (2026).
32. Wang, Y.-Q. *et al.* Making waves: Emerging risks as regulatory mismatch meets automation in llm-agents like openclaw for water treatment and environmental infrastructure. *Water Research* **308**, 126862 (2027).
33. Liao, Z., Gu, X., Xie, J., Wang, X. & Chen, J. An integrated assessment of drainage system reconstruction based on a drainage network model. *Environ Sci Pollut Res* **26**, 26563–26576 (2019).
34. Tian, W. *et al.* Enhancing interpretability and generalizability of deep learning-based emulator in three-dimensional lake hydrodynamics using Koopman operator and transfer learning: Demonstrated on the example of lake Zurich. *Water Research* **249**, 120996 (2024).
35. Shinn, N., Cassano, F., Gopinath, A., Narasimhan, K. & Yao, S. Reflexion: language agents with verbal reinforcement learning. in *Advances in Neural Information Processing Systems* (eds Oh, A. et al.) vol. 36 8634–8652 (Curran Associates, Inc., 2023).
36. Sarim, M., Masood, F., Maheshwari, M., Faridi, A. R. & Shamsan, A. H. Generating reliable software project task flows using large language models through prompt engineering and robust evaluation. *Sci Rep* **15**, 35194 (2025).

**From Rule-based to Agentic: LLM Agent based Real-time Control of Urban Drainage Systems to Mitigate Flooding and Sewer Overflow**

Wenchong Tian[a,c], Shixun Li[a,c], Ruozhou Lin[a,c], Peng Ye[d], Zhiyu Zhang[a,c], Shi Tao[a], Yu-Qi Wang[a,c], Ling Shi[e], Mingfu Guan[f], Haoran Duan[f]*, Zhiguo Yuan[a,b]*

[a] *School of Energy and Environment, City University of Hong Kong, Hong Kong SAR, China.*

[b] *State Key Laboratory of Marine Pollution, City University of Hong Kong, Hong Kong SAR, China.*

[c] *Shenzhen Research Institute, City University of Hong Kong, China.*

[d] *The Chinese University of Hong Kong, Hong Kong SAR, China.*

[e] *Department of Electronic and Computer Engineering, The Hong Kong University of Science and Technology, Hong Kong SAR, China.*

[f] *Department of Civil Engineering, The University of Hong Kong, Hong Kong SAR, China.*

* Corresponding authors

E-mail address:

Prof. Zhiguo Yuan (zhigyuan@cityu.edu.hk);

Prof. Haoran Duan (haoran.duan@hku.hk)

**S1. Prompt templates of Multi-modal Meteorological Monitor**

The Multi-modal Meteorological Monitor consists of two core components. The first component leverages LLMs to construct Python scripts for parsing rainfall datasets and meteorological warnings, and convert these warnings into standardized analytical outputs. This component runs in a one-off execution mode within this research to acquire full warning event records covering a specified time window. The second component is the real-time execution variant for meteorological warning monitoring. Configured to trigger execution every 10 minutes, it is designed for real-time invocation in practical engineering deployments.

Prompts for all components are listed below. These prompts were iteratively refined through continuous dialogue with AI (WorkBuddy with GLM-5.2), thereby incorporating abundant detailed specifications. Since the research scope is confined to mainland China, and target rainfall & warning data are crawled from Chinese-language web platforms, Chinese text is embedded within these prompt contents. The prompt Markdown files, together with the intermediate generated data reading scripts and data parsing scripts, are available on GitHub (https://github.com/DantEzio/LLMAgent_UDSControl).

(1) Prompt for rainfall data acquisition (Prompt_Rainfall_Get.md)

Please write a Python script to fetch hourly precipitation data for the Chaohu (巢湖) region from the Open-Meteo Historical Weather Archive API, and save the results as an Excel file.

## API Information

- API Endpoint: https://archive-api.open-meteo.com/v1/archive
- This is a free historical weather data API provided by Open-Meteo, based on ERA5 reanalysis data (ECMWF's global reanalysis dataset), with a spatial resolution of approximately 0.25° (~25 km grid). No API key is required.
- Official documentation: https://open-meteo.com/en/docs/historical-weather-api

## Request Parameters

- Latitude: 31.6 (Chaohu city center area)
- Longitude: 117.87 (Chaohu city center area)
- Start date: 2015-01-01
- End date: 2025-06-10
- Hourly variable: precipitation (precipitation amount in mm)
- Timezone: Asia/Shanghai (UTC+8, ensures timestamps are consistent with Beijing time)

## Output Requirements

Save the retrieved data as an Excel file named `Chaohu_Rainfall_2015-2025_OpenMeteo.xlsx`, containing the following 3 worksheets:

### Sheet 1: Hourly_Rainfall (Hourly Precipitation Data)

Each row represents one hour of precipitation record, with the following column structure:

| Column | Description | Example |
|--------|-------------|---------|
| datetime | Full date and time (Asia/Shanghai timezone) | 2015-01-01 00:00:00 |
| date | Date portion | 2015-01-01 |
| time | Time portion | 00:00 |
| year | Year | 2015 |
| month | Month | 1 |
| day | Day | 1 |
| hour | Hour | 0 |
| precipitation_mm | Precipitation amount (millimeters) | 0.0 |

Total rows should be approximately 91,536 (2015-01-01 00:00 to 2025-06-10 23:00, ~10.5 years × 365.25 days × 24 hours).

### Sheet 2: Yearly_Stats (Annual Statistics)

Aggregate precipitation data by year, with the following columns:

| Column | Description |
|--------|-------------|
| year | Year |
| total_precipitation_mm | Annual total precipitation (mm) |
| max_hourly_precipitation_mm | Annual maximum hourly precipitation (mm) |
| rainy_hours | Number of hours with precipitation (precipitation > 0) |
| rainy_days | Number of days with precipitation (daily total > 0) |

### Sheet 3: Monthly_Stats (Monthly Statistics)

Aggregate precipitation data by year-month, with the following columns:

| Column | Description |
|--------|-------------|
| year | Year |
| month | Month |
| total_precipitation_mm | Monthly total precipitation (mm) |
| max_hourly_precipitation_mm | Monthly maximum hourly precipitation (mm) |
| rainy_hours | Number of hours with precipitation (precipitation > 0) |

Total rows should be approximately 126 (10.5 years × 12 months).

## Technical Requirements

1. Use `requests` library to call the API (GET request, parameters passed via query string)
2. Use `pandas` for data processing and organization
3. Use `openpyxl` or `pandas.ExcelWriter` to save the Excel file
4. Implement a request retry mechanism (at least 3 retries with 5-second intervals) to handle network timeouts and API rate limiting
5. After receiving the response, print the JSON structure keys to confirm data integrity
6. Check the precipitation column for NaN values; if any are found, fill them with 0

7. After saving, read the Excel file back and verify that the row counts of all three sheets are correct

8. Print a final summary: total rows, time range, total precipitation, maximum hourly precipitation, number of rainy hours

## Notes

- ERA5 reanalysis data is gridded data, not station observation data. The coordinates 31.6°N, 117.87°E will be mapped to the nearest grid cell

- The precipitation variable is measured in millimeters (mm), representing the precipitation amount for that hour

- The API returns time strings in ISO 8601 format (e.g., "2015-01-01T00:00"), which need to be parsed into datetime objects

- When timezone=Asia/Shanghai is set, the API returns timestamps already in UTC+8; no additional conversion is needed

- The data volume is large (~91,000 rows) but can be retrieved in a single request. If a timeout occurs, consider splitting the request by year and merging the results afterward

(2) Prompt for acquiring meteorological warnings (Prompt_Warning_Get.md)

Please perform a systematic web search to collect all rainfall-related meteorological warnings issued by the National Meteorological Center (NMC) / China Meteorological Administration (CMA) that cover the Chaohu (巢湖) region from 2015 to 2025. Compile the results into a structured Markdown summary file and a folder of individual event directories.

## Target Region
- **Primary target**: Chaohu city (巢湖市), Anhui Province, China
- **Coordinates**: latitude 31.6°N, longitude 117.87°E
- **Coverage criteria**: Any warning whose stated coverage area includes at least one of the following:
  - "巢湖" (directly named)
  - "安徽中部" (central Anhui — Chaohu is in central Anhui)
  - "安徽中南部" (south-central Anhui)
  - "安徽中北部" (north-central Anhui)
  - "江淮" or "江淮之间" (Jianghuai region — Chaohu sits in this area)
  - "苏皖" or "苏皖鄂" (Jiangsu-Anhui or Jiangsu-Anhui-Hubei combined warnings)
  - "长江中下游" (middle and lower Yangtze — Chaohu is in this zone)
  - "安徽全省" or "安徽大部" (all or most of Anhui)
  - "合肥" (Hefei — Chaohu is administratively under Hefei)

## Warning Types to Collect
Collect the following types of warnings:
1. **Rainstorm warnings** (暴雨预警) — all levels: Blue (Ⅳ), Yellow (Ⅲ), Orange (Ⅱ), Red (Ⅰ)
2. **Typhoon warnings** (台风预警) — when the typhoon causes rainfall affecting Chaohu (include only if rainfall is mentioned)
3. **Severe convective weather warnings** (强对流预警) — when short-duration heavy rain is involved
4. **Weather bulletins** (天气公报) — when they mention heavy rain covering the Chaohu area
5. **Emergency responses** (应急响应) — CMA's meteorological disaster emergency response levels related to rainstorms
6. **Season reviews / post-event analyses** (汛期盘点/复盘) — when they include Chaohu as a case study

## Rainfall Amount Standards (GB/T 28592-2012)
Include this reference in the output file header:
- Blue warning (暴雨蓝色): 12h ≥ 50mm / 24h 50–100mm
- Yellow warning (暴雨黄色): 6h ≥ 50mm / 24h 100–250mm (heavy rainstorm)
- Orange warning (暴雨橙色): 3h ≥ 50mm / 24h 100–250mm (heavy rainstorm), locally ≥ 250mm (extraordinary)
- Red warning (暴雨红色): 3h ≥ 100mm / 24h ≥ 250mm (extraordinary rainstorm)

## Search Strategy
Search year by year (2015, 2016, ..., 2025), for each year:
1. **Search query**: Use web search with queries like:
   - `中央气象台 暴雨预警 巢湖 [year]`
   - `中国气象局 暴雨预警 安徽 [year]`
   - `CMA rainstorm warning Anhui Chaohu [year]`
   - `中央气象台 台风预警 安徽 [year]`
   - For each found warning, search for the specific date to find the original CMA page
2. **Primary data sources** (in priority order):
   - CMA official website: `cma.gov.cn` (China Meteorological Administration) — look for URLs matching pattern `cma.gov.cn/2011xwzx/...` or `cma.gov.cn/2011zwxx/...`
   - NMC official website: `nmc.cn` (National Meteorological Center)
   - CCTV news: `tv.cctv.com` or `news.cctv.com`
   - China Weather Net: `weather.com.cn`
   - National Emergency Broadcasting: `cneb.gov.cn`
   - Chaohu city government: `chaohu.gov.cn` (for local-level warnings)
   - Other credible media: The Paper (thepaper.cn), Xinhua, China News, CRI Online, CNR (cnr.cn)
3. **For each warning found, extract**:
   - **Issue date/time** (发布时间 — as precise as possible; if only date is found, note "(assumed)" for the time)
   - **Warning type and level** (预警类型 — e.g., "Rainstorm Orange Warning", "Typhoon Red Warning")
   - **Coverage area** (覆盖区域 — the original text describing which regions are covered)
   - **Predicted rainfall** (预报雨量 — extract the specific numbers from the NMC original text. If exact numbers are not available, infer the range from the warning level standard and mark it as "[Standard]" vs "[NMC Original]" for data traceability)
   - **Source** (来源 — the name of the publishing media, e.g., "CMA", "CCTV", "China Weather Net")
   - **URL** (网址 — the full link to the source page)
4. **Rainfall data labeling convention**:
   - `[NMC Original]` — Exact millimeter values extracted from the CMA/NMC original warning page
   - `[Standard]` — Range inferred from the warning level per GB/T 28592-2012
   - `[Provincial/City Station]` — Values from provincial or city-level meteorological stations
   - `[Continued Warning]` — Inferred from the previous day's NMC values when a warning is continued
   - `[CMA Review]` — Values from CMA post-event analysis reports

## Output File 1: Main Summary
Save as: `中央气象台_巢湖降雨预警_2015_2025.md`

### Structure:
```
```

# Summary of Rainfall Warnings for Chaohu Region by NMC/CMA (2015—2025)
> **Data sources:** National Meteorological Center (NMC) / China Meteorological Administration (CMA) official websites, CCTV news relay
> **Coverage area:** Chaohu city (including Chaohu basin, Hefei, Anhui Jianghuai region and extended coverage)
> **Warning types:** Rainstorm warnings (Blue/Yellow/Orange/Red), typhoon-rainfall warnings, severe convective weather warnings, weather bulletins
> **Compiled date:** [today's date]
> **Rainfall standard reference:** [include the GB/T 28592-2012 standards listed above]
---

## Table of Contents
1. [2015](#2015)
...
11. [2025](#2025)
12. [Summary Table](#summary-table)
13. [Statistical Overview](#statistical-overview)
---

## [Year]
[Brief paragraph: 1-3 sentences summarizing the year's rainfall pattern and major events for Chaohu]

| # | Issue Date/Time | Warning Type | Coverage Area | **Predicted Rainfall (NMC)** | Source | URL |
|:---:|------|------|------|------|------|------|
| 1 | 2015-06-02 | Rainstorm Blue | Central and southern Anhui (incl. Chaohu basin) | Heavy rain to rainstorm 25-100mm; **local heavy rainstorm 100-160mm** [NMC Original] | CMA | [link](URL) |
| ... | ... | ... | ... | ... | ... | ... |

> **[Year] summary:** [Summary: count, dominant warning level, key events, notable rainfall amounts]
---
[Repeat for each year 2015-2025]
---

## Summary Table
### All entries (sorted by year + date, with predicted rainfall)

| # | Year | Date | Warning Type | Chaohu Association | **Predicted Rainfall (core values)** | Source | URL |
|:---:|:---:|------|------|------|------|------|------|
| 1 | 2015 | 06-02 | Rainstorm Blue | Southern Anhui | Local **heavy rainstorm 100-160mm** | CMA | [link](URL) |
| ... | ... | ... | ... | ... | ... | ... | ... |

---

## Statistical Overview
### Statistics by Year

| Year | Count | Highest Warning Level | Main Causative Factor | Typical Chaohu Event | **Strongest Predicted Rainfall** |
|:---:|:---:|:---:|------|------|------|
| 2015 | 5 | Rainstorm Blue | Meiyu-front rainstorm | Late Jun sustained rain in Jiangsu-Anhui | **Heavy rainstorm 100-160mm** |
| ... | ... | ... | ... | ... | ... |

### Statistics by Predicted Rainfall Level

| Rainfall Level | Occurrences | Typical Events (Year) |
|:---|:---:|------|
| Heavy rain to rainstorm (25-100mm) | XX | ... |
| Rainstorm to heavy rainstorm (50-200mm) | XX | ... |
| Heavy rainstorm (100-250mm) | XX | ... |
| Extraordinary rainstorm (≥250mm) | XX | ... |

### Statistics by Source

| Source | Count | Percentage |
|:---|---:|---:|
| CMA/NMC official | XX | XX% |
| CCTV/Xinhua | XX | XX% |
| ... | ... | ... |

### Predicted Rainfall Data Source Legend

| Label | Meaning | Count |
|:---|------|:---:|
| [NMC Original] | Exact values from CMA/NMC original warning page | XX |
| [Standard] | Range inferred from warning level per GB/T 28592-2012 | XX |
| [Provincial/City Station] | Values from provincial/city meteorological stations | XX |
| [Continued Warning] | Inferred from previous day's NMC values | XX |
| [CMA Review] | Values from CMA post-event analysis | XX |

### Key Findings

[5-6 bullet points summarizing key findings: extreme years, top rainfall forecasts, direct Chaohu mentions, typhoon contributions, warning frequency trends, data traceability rate]

```

## Output File 2: Event Summary Files

In the `NMC_Events_2015_2025/` folder, create:

### 2a. `00_Event_Summary.md`

A markdown summary table of all events with rainfall data statistics:

```
# NMC Warning Events Summary (2015-2025)
**Total events:** XX
**Events with rainfall data:** XX (XX%)
**Time window per event:** 72 hours = 24h before + 24h rainfall + 24h after warning
**Warning time assumption:** 08:00 on event date

## All Events
| # | Date | Warning Type | Location | Total Rain (mm) | Max Intensity (mm/h) | Rain Hours | Records | Data |
|:---:|------|------|------|:---:|:---:|:---:|:---:|:---:|
| 01 | 2015-06-02 | Rainstorm Blue | Southern Anhui (incl. Chaohu basin) | 69.1 | 7.2 | 31 | 72 | ✅ |
| ... | ... | ... | ... | ... | ... | ... | ... | ... |
```

### 2b. `00_Event_Summary.csv`

Same data as above in CSV format with columns:

`seq, date, warning_type, location, total_rainfall_mm, max_intensity_mmh, rain_hours, folder, n_records`

### 2c. Individual Event Folders
```

For each event, create a folder named:
```
Event_[XX]_[YYYY-MM-DD]_[Warning_Type_Abbreviation]
```
Examples: `Event_01_2015-06-02_暴雨蓝色`, `Event_06_2016-06-30_暴雨橙色_2016 年首个`
Each event folder contains the following files:
#### README.md
```
# Event [XX]: [YYYY-MM-DD] [Warning Type]

## Event Information
| Item | Content |
|:---|------|
| **Event number** | XX |
| **Warning issue time** | YYYY-MM-DD HH:MM |
| **Warning type** | [full warning type] |
| **Coverage area** | [coverage area text] |
| **Predicted rainfall** | [predicted rainfall] |
| **Data source** | CMA / CCTV / ... |
| **Original URL** | [source URL if available] |

## Rainfall Data Summary
| Metric | Value |
|:---|:---:|
| **Data window** | [start] ~ [end] (72 hours) |
| **Total rainfall** | XX.X mm |
| **Max hourly intensity** | XX.X mm/hr |
| **Rain hours** | XX / 72 |

## Time-Period Statistics
| Period | Time Range | Cumulative Rainfall | Max Intensity |
|:---|------|:---:|:---:|
| Pre-warning 24h | [range] | XX.X mm | XX.X mm/hr |
| Rainfall process 24h | [range] | XX.X mm | XX.X mm/hr |
| Post-warning 24h | [range] | XX.X mm | XX.X mm/hr |

## File List
- `Event_XX_..._rainfall.csv` — Hourly rainfall data
- `Event_XX_..._SWMM.npy` — SWMM format rainfall series (72h, shape=[72,2])
- `Event_XX_warnings.md` — Detailed warning records (if available)
```
#### [EventName]_rainfall.csv
72-hour hourly rainfall data extracted from the Open-Meteo ERA5 Excel file. CSV with header:
```
datetime,date,time,year,month,day,hour,precipitation_mm
```

Time window: 24h before warning time + 24h of rainfall process + 24h after = 72 rows total.
Warning time assumed as 08:00 on the event date if exact time is unknown.
#### [EventName]_SWMM.npy
NumPy array, shape=[72, 2], columns: [elapsed_hours, rainfall_intensity_mmh]. This is the SWMM (Storm Water Management Model) format rainfall input time series.
#### [EventName]_warnings.md (for early events 2015-2016, if CMA archive pages are accessible)
Detailed warning records within the rainfall time window, including:
- Time window definition: rain-before-6h + rainfall-duration + rain-after-48h
- Each warning record as a table with: issue time, issuing agency, warning level, coverage area, rainfall prediction, max hourly intensity, remarks
- Warning evolution timeline (text diagram showing level changes over time)
- Local warning supplements (provincial/city-level warnings)

## Notes
- Some early warnings (2015-2017) may have their original CMA pages archived or removed. Use CCTV/Xinhua relay reports and infer rainfall amounts from the warning level standard, marking them with the appropriate data label tag
- The "predicted rainfall" is the **forecast** value from NMC's original warning (not observed/actual rainfall). Actual observed values may be mentioned in parentheses for comparison when available
- Events beyond the rainfall data coverage period (after 2025-06-10) will have no rainfall CSV/npy files, only the README.md — mark them with "N/A" and "❌" in the summary
- Folder naming: replace spaces with underscores, keep Chinese characters for warning type abbreviations
- Each event should have a sequential number (01, 02, ..., 52) assigned in chronological order
- When a single date has multiple warnings (e.g., typhoon orange + rainstorm orange), create separate event folders for each

(3) Prompt for retrieving meteorological warnings at a 10-minute interval (Prompt_Realtime_Warning_Monitoring.md)

Please write a Python script that polls the National Meteorological Center (NMC) / China Meteorological Administration (CMA) website every 10 minutes for currently active rainfall-related warnings covering the Chaohu region. When new or updated warnings are found, save them to the `10min_warnings/` folder. This script is designed to be executed every 10 minutes via an external scheduler (e.g., Windows Task Scheduler or cron).

## Target Region
- **Primary target**: Chaohu city (巢湖市), Anhui Province, China
- **Coordinates**: latitude 31.6°N, longitude 117.87°E
- **Coverage criteria**: Any warning whose stated coverage area includes at least one of the following:
  - "巢湖" (directly named)
  - "安徽中部" (central Anhui) / "安徽中南部" (south-central Anhui) / "安徽中北部" (north-central Anhui)
  - "江淮" or "江淮之间" (Jianghuai region — Chaohu sits in this area)
  - "苏皖" or "苏皖鄂" (Jiangsu-Anhui or Jiangsu-Anhui-Hubei combined warnings)
  - "长江中下游" (middle and lower Yangtze — Chaohu is in this zone)
  - "安徽全省" or "安徽大部" (all or most of Anhui)
  - "合肥" (Hefei — Chaohu is administratively under Hefei)

## Warning Types to Monitor
1. **Rainstorm warnings** (暴雨预警) — Blue / Yellow / Orange / Red
2. **Typhoon warnings** (台风预警) — when rainfall is mentioned
3. **Severe convective weather warnings** (强对流预警)
4. **Weather bulletins** (天气公报) — when heavy rain covers the Chaohu area

## Data Sources (Real-Time Warning Pages)
The script should fetch and parse the following pages in priority order:
### Primary: NMC Official Warning Page
- URL: `http://www.nmc.cn/publish/country/warning/index.html`
- This page lists all currently active national-level meteorological warnings
- Parse the HTML to extract warning entries: warning type, level, coverage area, issue time, and any detail link
### Secondary: CMA Warning Release Page
- URL: `https://www.cma.gov.cn/2011xwzx/2011sqyj/`
- CMA's official severe weather warning release page
- Contains the latest warnings issued by CMA headquarters
### Tertiary: China Weather Net Warning Page
- URL: `http://www.weather.com.cn/alarm/index.shtml`
- Alternative real-time warning feed
### Fallback: Anhui Provincial Meteorological Bureau
- URL: `https://www.ahqx.cn/` (Anhui Provincial Meteorological Bureau)
- Provincial-level warnings that may include Chaohu-specific alerts

## Request Parameters
- Use `requests` with a realistic User-Agent header (e.g., `"Mozilla/5.0 (Windows NT 10.0; Win64; x64) AppleWebKit/537.36"`)
- Set timeout to 15 seconds per request
- Implement retry: 3 retries with 5-second intervals on failure
- If all data sources fail in a single run, still record the attempt with a "FAILED" status (do not crash silently)

## Output Requirements
### Folder Structure
All output files are saved in the `10min_warnings/` directory (create it if it does not exist).
### File Format: Daily JSON Lines + Daily CSV
Each day generates two files plus a persistent state file:
```
10min_warnings/
  chaohu_warnings_2025-07-29.jsonl      # One JSON object per line (full detail)
  chaohu_warnings_2025-07-29.csv        # Flat table for quick inspection
  _dedup_state.json                     # Persistent deduplication state
```
### JSONL Record Structure (one line per warning entry per poll)
```json
{
  "fetch_time": "2025-07-29 14:37:05",
```

```
    "source": "NMC",
    "source_url": "http://www.nmc.cn/publish/country/warning/index.html",
    "warning_id": "202507291400-暴雨橙色-安徽中南部",
    "issue_time": "2025-07-29T14:00",
    "warning_type": "暴雨",
    "warning_level": "橙色",
    "coverage_area": "安徽中南部（含巢湖流域）",
    "predicted_rainfall": "3h 50-100mm",
    "detail_url": "http://www.nmc.cn/publish/country/warning/14xxxx.html",
    "chaohu_match_reason": "安徽中南部",
    "status": "ACTIVE",
    "is_new": true,
    "is_updated": false
}
```

### CSV Column Structure

| Column | Description | Example |
|--------|-------------|---------|
| fetch_time | System timestamp when script runs | 2025-07-29 14:37:05 |
| source | Data source name | NMC |
| warning_id | Dedup identifier (issue_time + type + level + area) | 202507291400-暴雨橙色-安徽中南部 |
| issue_time | Warning issue time from source | 2025-07-29T14:00 |
| warning_type | Type of warning | 暴雨 |
| warning_level | Level (Blue/Yellow/Orange/Red) | 橙色 |
| coverage_area | Original coverage text | 安徽中南部（含巢湖流域） |
| predicted_rainfall | Predicted rainfall text | 3h 50-100mm |
| chaohu_match_reason | Which keyword matched | 安徽中南部 |
| detail_url | Link to warning detail page | http://... |
| status | ACTIVE / EXPIRED / FAILED | ACTIVE |
| is_new | First time seeing this warning? | True |
| is_updated | Warning level/content changed since last seen? | False |

If today's files do not exist yet (first run of the day), create them with a header row (CSV). If they already exist, append new rows without rewriting the header.

### File Encoding

CSV: UTF-8 with BOM (utf-8-sig) for Excel compatibility on Windows.

JSONL: UTF-8 (no BOM).

## Deduplication Mechanism

This is the most critical component — the script runs every 10 minutes, so the same warning will appear in multiple polls. To avoid saving duplicates:

### State File: `_dedup_state.json`

Stored in the `10min_warnings/` folder. Structure:

```json
{
    "active_warnings": {
        "202507291400-暴雨橙色-安徽中南部": {
```

```
      "warning_id": "202507291400-暴雨橙色-安徽中南部",
      "first_seen": "2025-07-29 14:07:05",
      "last_seen": "2025-07-29 14:37:05",
      "warning_level": "橙色",
      "coverage_area": "安徽中南部（含巢湖流域）",
      "predicted_rainfall": "3h 50-100mm",
      "poll_count": 4
    }
  },
  "expired_warnings": {},
  "last_cleanup": "2025-07-29 14:37:05"
}
```

### Dedup Logic

1. **Build warning_id**: Combine `issue_time + warning_type + warning_level + coverage_area` into a unique identifier string
2. **On each poll**, compare fetched warnings against `active_warnings` in the state file:
   - **New warning** (`is_new: true`): warning_id not in active_warnings → save to JSONL/CSV, add to state
   - **Existing warning, no change** (`is_new: false, is_updated: false`): warning_id exists, content unchanged → update `last_seen` and `poll_count` in state, do NOT write to JSONL/CSV (or optionally write a lightweight "still active" heartbeat row)
   - **Updated warning** (`is_updated: true`): warning_id exists but level/coverage/rainfall changed → save to JSONL/CSV with `is_updated: true`, update state
3. **Expiry detection**: Warnings in `active_warnings` that are NOT found in the current poll → move to `expired_warnings` with `expired_at` timestamp, do NOT write to JSONL/CSV
4. **State cleanup**: Every 24 hours (or when `expired_warnings` exceeds 200 entries), remove entries older than 7 days from `expired_warnings`

### Warning ID Construction

```python
def build_warning_id(issue_time, warning_type, warning_level, coverage_area):
    # Normalize: remove spaces, punctuation, lowercase
    import hashlib
    raw = f"{issue_time}|{warning_type}|{warning_level}|{coverage_area}"
    return hashlib.md5(raw.encode('utf-8')).hexdigest()[:16]
```

Alternatively, use the human-readable format: `{issue_time}_{warning_type}{warning_level}_{coverage_area_truncated}`.

## Technical Requirements

1. Use `requests` with `BeautifulSoup` (bs4) for HTML parsing
2. Use the `csv` module for CSV appending (standard library, lightweight)
3. Use `json` module for JSONL writing and state file management
4. Use `datetime` module for timestamps and daily filenames
5. Implement request retry: 3 retries, 5-second intervals
6. If all data sources fail, append a row with `status: "FAILED"` and `warning_id: ""` — this makes data gaps visible
7. Print a concise log line to stdout after each run:

   ```
   ```

[2025-07-29 14:37:05] Source: NMC | Active warnings: 3 | New: 1 | Updated: 0 | Expired: 0 | Saved to 10min_warnings/chaohu_warnings_2025-07-29.csv
```

8. The script should be a single-run script (fetch once, process, save, then exit). Do NOT include a `while True` loop or `time.sleep()` — the 10-minute interval is handled by the external scheduler
9. Handle HTML encoding properly (NMC pages may use GB2312/GBK or UTF-8 — detect from response headers or content)
10. Be respectful to the server: do not send requests more frequently than the 10-minute interval allows

## HTML Parsing Notes
NMC warning pages may use JavaScript rendering. If `requests` + BeautifulSoup cannot extract warning data (empty results), consider:
- Using the `agent-browser` skill (Playwright headless browser) to render JavaScript first, then extract DOM
- Checking if NMC provides a JSON API behind the warning page (inspect network requests in browser DevTools)
- As a fallback, search for the warning RSS/Atom feed if available

## Scheduling Instructions (include as comments at the top of the script)
### Windows Task Scheduler
1. Open Task Scheduler → Create Basic Task
2. Set trigger: Daily, repeat every 10 minutes for 1 day
3. Set action: "Start a program" → point to Python executable with script path as argument
### Linux / macOS crontab
```
*/10 * * * * /path/to/python /path/to/fetch_warnings_10min.py >> /path/to/10min_warnings/cron.log 2>&1
```

## Rainfall Amount Standards (GB/T 28592-2012)
For reference when parsing predicted rainfall from warning text:
- Blue: 12h ≥ 50mm / 24h 50–100mm
- Yellow: 6h ≥ 50mm / 24h 100–250mm
- Orange: 3h ≥ 50mm / 24h 100–250mm
- Red: 3h ≥ 100mm / 24h ≥ 250mm

## Notes
- NMC warning pages are primarily in Chinese — parsing logic should handle Chinese text
- Some warnings may be issued at the national level (covering multiple provinces) while others are provincial — both should be checked for Chaohu keywords
- Warning level upgrades/downgrades (e.g., Blue → Yellow → Orange) are common during active weather events — the dedup mechanism handles this via `is_updated: true`
- The `active_warnings` state file should be backed up daily (copy to `10min_warnings/_archive/dedup_state_YYYY-MM-DD.json`)
- If the NMC website changes its HTML structure, the parsing selectors will need updating — include a clear comment block identifying all CSS selectors used for easy maintenance

(4) Prompt for fusing rainfall data and warning data (Prompt_Process_Events.md)
```

Write a Python script `process_events.py` that reads hourly rainfall data from an Excel file and a list of 52 meteorological warning events (hardcoded), then for each event extracts a 72-hour rainfall window and outputs three files per event (CSV, NPY, Markdown README), plus an overall summary (CSV + Markdown).

## Input Data
### Input 1: Excel File — `Chaohu_Rainfall_2015-2025_OpenMeteo.xlsx`
- **Location:** Same directory as the script (`BASE_DIR`).
- **Sheet name:** `Hourly_Rainfall`
- **Total rows:** ~91,536 (hourly records from 2015-01-01 00:00 to 2025-12-31 23:00)
- **Columns (8 columns):**
| Column | Dtype | Description |
|--------|-------|-------------|
| `datetime` | datetime64 | Full timestamp (e.g., `2015-06-27 08:00:00`) |
| `date` | str | Date string `YYYY-MM-DD` |
| `time` | str | Time string `HH:MM:SS` |
| `year` | int64 | Year (2015–2025) |
| `month` | int64 | Month (1–12) |
| `day` | int64 | Day of month |
| `hour` | int64 | Hour (0–23) |
| `precipitation_mm` | int64 | Hourly precipitation in millimeters |
- **Data source:** Open-Meteo ERA5 reanalysis, 0.25° grid, coordinates lat=31.6, lon=117.87 (Chaohu region).
- **Other sheets** (`Yearly_Stats`, `Monthly_Stats`) are NOT used by this script.
### Input 2: 52 Event Definitions — Hardcoded as Python List
The 52 events come from the summary table in `中央气象台_巢湖降雨预警_2015_2025.md`, but they are **NOT read programmatically from the markdown**. Instead, they are hardcoded as a Python list of tuples inside the script.

Each event tuple has **8 fields**:
```python
(seq, year, date_str, warning_type, location, forecast_rain, source, url)
```
| Field | Type | Example | Description |
|-------|------|---------|-------------|
| `seq` | int | `1` | Sequential number 1–52 |
| `year` | int | `2015` | Year of the event |
| `date_str` | str | `"06-02"` | Month-day string (MM-DD) |
| `warning_type` | str | `"暴雨蓝色预警"` | Full warning type text (Chinese) |
| `location` | str | `"安徽中南部(含巢湖流域)"` | Coverage area |
| `forecast_rain` | str | `"局地大暴雨 100-160mm"` | NMC forecast rainfall description |
| `source` | str | `"CMA"` | Data source abbreviation |
| `url` | str | `"https://www.cma.gov.cn/..."` | Original URL |

## Output Data
### Output Directory Structure
```
NMC_Events_2015_2025/
├── 00_Event_Summary.csv ← Overall summary (CSV)
```

```
├── 00_Event_Summary.md ← Overall summary (Markdown)
├── Event_01_2015-06-02_暴雨蓝色/
│   ├── README.md ← Event metadata + rainfall statistics
│   ├── Event_01_2015-06-02_暴雨蓝色_rainfall.csv ← Full hourly rainfall data
│   └── Event_01_2015-06-02_暴雨蓝色_SWMM.npy ← SWMM format time series
├── Event_02_.../
│   └── ...
└── ... (52 event folders total)
```

### Output File 1: `{folder_name}_rainfall.csv` (per event)

- Contains all original columns from the Excel sheet (`datetime`, `date`, `time`, `year`, `month`, `day`, `hour`, `precipitation_mm`) filtered to the 72-hour window.
- Encoding: `utf-8-sig` (for Excel compatibility).
- No index column.

### Output File 2: `{folder_name}_SWMM.npy` (per event)

- **Format:** NumPy 2D array, `np.float64`.
- **Shape:** `(N, 2)` where N = number of hourly records in the window (typically 72).
- **Column 0:** Elapsed hours from window start (0, 1, 2, ..., N-1).
- **Column 1:** Precipitation intensity in mm/hour (from `precipitation_mm` column).
- **Usage:** Direct input to SWMM (Storm Water Management Model) rainfall time series.

```python
# Construction:
elapsed_hours = np.arange(len(event_df), dtype=np.float64)
precipitation = event_df['precipitation_mm'].values.astype(np.float64)
swmm_ts = np.column_stack([elapsed_hours, precipitation])
np.save(npy_path, swmm_ts)
```

### Output File 3: `README.md` (per event)

Markdown document with three sections:

1. **事件信息 (Event Info)** — Table with: 事件序号，预警发布时间，预警类型，覆盖区域，预报雨量，数据来源，原始网址
2. **降雨数据摘要 (Rainfall Summary)** — Table with: 数据窗口 (72h range)，总降雨量，最大小时雨强 (with timestamp)，降雨小时数 (count/total)
3. **分时段统计 (Segment Statistics)** — Table splitting the 72h window into three 24h segments: 预警前 24h，降雨过程 24h，预警后 24h, each with 累计降雨量 and 最大雨强
4. **文件列表** — Lists the two data files with descriptions

### Output File 4: `00_Event_Summary.csv` (overall)

Columns: `seq`, `date`, `warning_type`, `location`, `total_rainfall_mm`, `max_intensity_mmh`, `rain_hours`, `folder`, `n_records`

### Output File 5: `00_Event_Summary.md` (overall)

Markdown document with:

- Total events processed count + error count
- Summary table (all 52 events in one table)
- SWMM Time Series Format explanation
- File Structure tree diagram

---

## Processing Logic

### Step 1: Load Excel Data

```python
df = pd.read_excel(EXCEL_PATH, sheet_name='Hourly_Rainfall')
df['datetime'] = pd.to_datetime(df['datetime'])
```

### Step 2: For Each Event, Extract 72-Hour Window

```
Warning time: assumed as 08:00 AM on the event date
  warning_dt = Timestamp(year=year, month=month, day=day, hour=8, minute=0)
Window definition (72 hours total):
  start_dt = warning_dt - 24 hours    (24h before warning)
  end_dt     = warning_dt + 48 hours    (24h rainfall + 24h after)    [exclusive]
Filter: df[(df['datetime'] >= start_dt) & (df['datetime'] < end_dt)]
```

### Step 3: Folder Naming Convention

```
wtype_short = warning_type with "预警" removed, parentheses replaced by "_", spaces replaced by "_"
              truncated to 30 chars max
loc_short     = location text before first "(" or "（", stripped truncated to 20 chars max
folder_name = "Event_{seq:02d}_{year}-{date_str}_{wtype_short}"
```

Example: `Event_03_2015-06-27_暴雨蓝色`

### Step 4: Generate Three Output Files

For each event:

1. Save CSV with `to_csv(encoding='utf-8-sig', index=False)`
2. Build and save NPY with `np.save()`
3. Build and save README.md with event metadata + rainfall statistics

### Step 5: Segment Statistics (for README.md)

Split the 72-hour window into three segments:

- `before_df`: `datetime < warning_dt` (24h before warning)
- `during_df`: `warning_dt <= datetime < warning_dt + 24h` (24h during rainfall)
- `after_df`: `datetime >= warning_dt + 24h` (24h after)

Each segment reports: `sum(precipitation_mm)` and `max(precipitation_mm)`.

### Step 6: Overall Summary

After all events, create:

- `00_Event_Summary.csv` from a list of per-event summary dicts
- `00_Event_Summary.md` with table + SWMM format explanation + file structure tree

### Error Handling

- If no data in the 72h window (e.g., date out of Excel range), print error and skip that event.
- Wrap each event processing in try/except, count successes and failures.

---

## Technical Requirements

- **Language:** Python 3.x
- **Libraries:** `pandas`, `numpy`, `os`, `sys`

- **Path handling:** Use `os.path.join()` for cross-platform compatibility
- **Encoding:** All CSV files use `utf-8-sig`; all Markdown files use `utf-8`
- **Constants:**
  ```python
  BASE_DIR = r"C:\Users\chong\Desktop\DRL_Agent_UDSRTC\Scenarios"
  EXCEL_PATH = os.path.join(BASE_DIR, "Chaohu_Rainfall_2015-2025_OpenMeteo.xlsx")
  EVENTS_DIR = os.path.join(BASE_DIR, "NMC_Events_2015_2025")
  ```
- **Progress output:** Print progress for each event with `[seq] date type: N records, total=Xmm, max=Ymm/h → folder_name`
- **Final output:** Print total success/error count and output directory path

---

## Context

This script is part of a AI agent project for urban drainage system real-time control (UDSRTC). The 52 meteorological warning events cover 2015–2025 for the Chaohu (巢湖) region in Anhui Province, China. The extracted 72-hour rainfall windows serve as simulation scenarios for SWMM-based drainage control.

The warning events include: torrential rain warnings (blue/yellow/orange/red levels per GB/T 28592-2012), typhoon-related rainfall warnings, and emergency response activations issued by the National Meteorological Center (NMC) / China Meteorological Administration (CMA) that cover the Chaohu watershed.

## S2. Prompt templates of Control Guide

The Control Guide consists of four core components. the formulation of tailored control recommendations, and a decision-gating that evaluates the feasibility of the proposed adjustments. The first component embeds prior domain constraints from UDS RTC practice to confine the agentic AI's output to a well-defined decision scope: whether to activate pumps for drainage in advance, and which specific pumps to engage based on spatial analysis and on-site system control protocols. The second component is established to determine whether adopting the agentic AI's recommendations is opportune at each control timestep.

The specific prompts are given below. These prompts were iteratively refined through continuous dialogue and vibe coding with AI (CodeBuddy with GLM-5.2), thereby incorporating abundant detailed specifications with diversified information. Since these prompt templates need to adapt to different operating conditions, they contain pending parameters that must be populated according to real-time operating conditions. We distinguish these parameters in the prompts using red color. Since the case study is located in China, the prompts in the open-source code contain Chinese text. The English version of these prompts is provided below.

(1) Prompt for control recommendations (given in llm_action_advisor.py)

SYSTEM_PROMPT = """

You are an expert in real-time control for the Chaohu urban drainage system.

Your task is to propose correction suggestions for pump control actions output by the PPO reinforcement learning model, based on rainfall early warnings, real-time monitoring data, and the BC control rules.

Core Principles:

1. The purpose of Series S emptying pumps (CC-S1/S2, JK-S) is to pre-emptively drain storage tanks and reserve capacity; they must be activated within a reasonable time window.

2. Operation strategy for emptying pumps: Series S emptying pumps (CC-S1/S2, JK-S) can run moderately under normal conditions; gradually ramp up operation within 1 hour before forecast rainfall; maintain continuous operation and prioritize activation if heavier rainfall is predicted.

3. If the predicted rainfall start time is more than 1 hour away, pre-emptive emptying shall not be initiated temporarily.

4. Flood & storage pumps (Series R: CC-R1/R2, JK-R2) can be activated when heavy rainfall is forecast and storage tank water level is non-zero; corresponding Series S emptying pumps must be turned on at the same time.

5. Gradual Intervention Principle: Select appropriate operation levels according to emergency severity to avoid over-aggressive adjustments:

- Level 0: No intervention (no early warning / too early timing / low water level)
- Level 1: Mild pre-emptive drainage — only activate CC-S1 (primary Chengguan emptying pump)
- Level 2: Moderate pre-emptive drainage — activate CC-S1 + JK-S (two main emptying pumps)

- Level 3: Aggressive pre-emptive drainage — activate all Series S pumps (CC-S1, CC-S2, JK-S)
Reference factors for level selection: Time to rainfall onset, storage tank water level, rainfall intensity trend
Default to lower levels (Level 1 or Level 2); only use Level 3 when emergency status is confirmed.

6. R Pump Constraint: Keep Series R pumps (CC-R1/R2, JK-R2) unchanged from the original BC outputs; only adjust Series S pumps (index 0, 1, 4).
7. Emergency Principle: When storage tanks have high water levels or severe flooding & CSO occurs, upgrade intervention to Level 3 or consider enabling R pumps.
8. Avoid premature drainage which causes energy waste and unnecessary pipeline pressure fluctuations.
9. Balance flood control safety and operational economy.
10. Preference Rule: When balancing flood safety and operating costs, prioritize pump operation; accept moderate energy consumption to avoid flood risks.
11. Action Constraint: You may only revise pumps from OFF (0) to ON (1) in original PPO actions. You are strictly prohibited from changing any pump state from ON (1) to OFF (0). For every pump i, adjusted_action[i] >= original_action[i] must hold true.

Only output JSON format, no extra text.
"""

```
FINAL_PROMPT =
"""You are a real-time control specialist for the Chaohu urban drainage system. Based on the information provided below, propose revision suggestions for the existing pump station control actions.
=== Rainfall Early Warning Information ===
{readme_snippet}
=== Event Analysis Results ===
Event Type: {warning_type}
Event Location: {location}
Estimated Total Rainfall: {total_rain:.1f} mm
Peak Rainfall Intensity: {peak_intensity:.1f} mm/hr
Duration: {duration} h
=== Monitored Real-Time System Status ===
{realtime_state_str}
{storage_eval_str}
=== BC Control Rules ===
{BC_RULES.strip()}
=== Pipe Network Connectivity Topology ===
{topo_summary}
=== Timing & Emptying Strategy Guidelines ===
{timing_guidance_str}
=== Original Actions ===
7 pump stations: {self.pump_names}
Raw PPO Actions (0 = Closed, 1 = Open): {at_original}
=== Output Revised Actions in JSON Format ===
Output only a valid JSON object strictly formatted as shown below:
{{
    "reasoning": "Brief explanation for revision rationale",
    "adjusted_action": [0 or 1, 0 or 1, 0 or 1, 0 or 1, 0 or 1, 0 or 1, 0 or 1],
```

```
    "urgency": "high/medium/low",
    "operation_level": Integer ranging from 0 to 3
}}
Definition of Operation Levels:
- Level 0: No intervention (retain original PPO actions)
- Level 1: Mild pre-emptive emptying (activate CC-S1 only)
- Level 2: Moderate pre-emptive emptying (activate CC-S1 + JK-S)
- Level 3: Aggressive pre-emptive emptying (activate all S-series pumps)
Notes: The sequence of values in adjusted_action must match the order of pump_names. No extra content shall be output.
【Critical Constraint on Action Modification】You are only allowed to switch pumps from closed state (0) to open state (1). It is forbidden to turn an open pump (1) to closed state (0). In other words, each element in adjusted_action must be greater than or equal to the corresponding element in the raw PPO action array.
【Critical Constraint on Progressive Strategy】Prioritize Level 1 or Level 2 for pre-emptive emptying. Level 3 shall only be adopted under emergency scenarios, including: lead time to rainfall ≤ 30 minutes, rainfall has commenced, storage occupancy ≥ 90%, or severe flooding occurs. The operating status of R-series pumps (CC-R1/R2, JK-R1/JK-R2) must remain identical to the original PPO actions.
"""


BC_RULES = """
BC Control Rules (Based on Storage Tank Water Level):
- CC-storage: Activate CC-S1/S2 pumps to empty and reserve storage capacity when water level > 3.0m; activate all CC pumps when water level > 4.0m
- JK-storage: Activate JK-S pump to empty and reserve storage capacity when water level > 4.5m; activate all JK pumps when water level > 5.5m
- Flood & Storage Pumps (CC-R1/R2, JK-R2): Can be turned on if heavy rainfall is forecast and the storage tank has non-zero water volume; corresponding emptying pumps (Series S) must be activated simultaneously when starting these pumps
- Operation strategy for emptying pumps: Series S emptying pumps (CC-S1/S2, JK-S) can run moderately under normal conditions; gradually ramp up operation within 1 hour before forecast rainfall; maintain continuous operation and prioritize activation if heavier rainfall is predicted
- Intensity Principle: Once pre-emptive emptying pumps are activated, operate them at maximum capacity as much as possible
- Emergency Principle: When storage tanks have high water levels or severe flooding & combined sewer overflow (CSO) occurs, both flood pumps and emptying pumps shall be operated to accelerate drainage
"""


timing_guidance_str = ""
if current_sim_time is not None:
    time_str = current_sim_time.strftime('%Y-%m-%d %H:%M')
    timing_guidance_str += f"Current simulation time: {time_str}\n"
    if warning_record is not None:
        pub_time = getattr(warning_record, 'publish_time', None)
        if pub_time:
```

```
            timing_guidance_str += f"Warning issuance time: {pub_time}\n"
        level = getattr(warning_record, 'level', None)
        if level:
            timing_guidance_str += f"Warning level: {level.value}\n"
    if rainfall_est is not None:
        timing_guidance_str += (
            f"Estimated total rainfall: {total_rain:.1f}mm\n"
            f"Estimated rainfall duration: {duration}h\n"
            f"Peak rainfall intensity: {peak_intensity:.1f}mm/hr\n"
        )
    timing_guidance_str += (
        "\n【Key Pre-Drainage Strategy Principles】\n"
        "1. The purpose of S-series draining pumps (CC-S1/S2, JK-S) is to pre-drain and reserve storage tank capacity, which must be activated within a reasonable time window.\n"
        "2. If it is more than one hour before the predicted rainfall starts, high-intensity pre-drainage shall not be initiated temporarily.\n"
        "3. Once pre-drainage pumps are activated, operate them at the maximum feasible output as much as possible.\n"
        "4. S-series draining pumps (CC-S1/S2, JK-S) can run normally under regular conditions; gradually start them within one hour before forecast rainfall. If heavier rainfall is expected, keep them running continuously with priority activation.\n"
        "5. R-series flood/storage pumps (CC-R1/R2, JK-R1/R2) can be activated when heavy rain is predicted and the storage tank water level is non-zero; corresponding S-series draining pumps must be started simultaneously when turning on R pumps.\n"
        "6. If the storage tank has a high water level or severe flooding and CSO occurs, both flood pumps and draining pumps need to operate to accelerate drainage.\n"
        "7. Avoid premature draining which leads to energy waste and unnecessary pipeline pressure fluctuations."
    )
else:
    timing_guidance_str = "No timing information available"
```

(2) Prompt for timing check (given in llm_action_advisor.py)

```
TIMING_VALIDATION_SYSTEM_PROMPT = """
You are a timing evaluation expert for the Chaohu urban drainage real-time control system.
Your task is to judge whether it is appropriate to implement the control adjustments recommended by the LLM at the current time.
Key Principles:
1. If more than one hour remains before the forecast rainfall onset and current rainfall intensity is zero → Timing is inappropriate (suitable=false)
2. If actual rainfall has started and shows an upward trend → Timing is appropriate (suitable=true)
3. If there is a clear rainfall warning within the one-hour window prior to rainfall → Gradual pump activation is allowed (suitable=true)
4. Operation rule for emptying pumps: CC-S1/S2 and JK-S emptying pumps can run at a moderate load under normal conditions; gradually start them within one hour before forecast rainfall. If heavier rainfall is predicted, maintain continuous
```

operation with priority activation.

5. Intensity Principle: Once a control adjustment is determined, operate pumps at the maximum feasible capacity.

6. Emergency Principle: If the storage tank holds a high water level or severe flooding/CSO occurs, all pump stations shall be activated to accelerate drainage.

7. Flood control safety and operational economy must be balanced to avoid unnecessary energy waste caused by premature emptying.

Only output JSON format: {"suitable": true/false, "reason": "..."}

"""

```
FINAL_PROMPT_P1 =
"""=== LLM Adjustment Strategy for Verification ===
LLM-suggested control actions: {list(llm_action) if llm_action else 'N/A'}
LLM reasoning statement: {llm_reasoning}
This strategy is formulated based on the following analytical data:
- Expected rainfall start time: {expected_rain_str}
"""
if est_rainfall is not None:
    prompt += (
        f"- Estimated total rainfall: {est_rainfall:.1f} mm\n"
        f"- Peak rainfall intensity: {peak_intensity:.1f} mm/hr\n"
        f"- Rainfall duration: {duration_hours} h\n"
    )

FINAL_PROMPT_P2 =
"""
=== Current Timing Information ===
Current simulation timestamp: {current_time_str}
Time remaining until expected rainfall onset: {time_until_rain_str}
Measured real-time rainfall intensity: {current_rainfall_value:.2f} mm/hr (a value of 0 indicates rainfall has not started)
"""
if storage_trend is not None:
    prompt += f"Storage tank water level trend: {storage_trend}\n"
"""
=== Judgment Requirements ===
```

Determine whether it is appropriate to implement the aforementioned LLM-recommended control strategy adjustment at the current moment according to the given information:

Judgment Rules:

1. If the lead time to predicted rainfall exceeds 1 hour and measured real-time rainfall equals zero → suitable=false (premature operation)

2. If rainfall has commenced with continuously rising rainfall intensity → suitable=true (appropriate timing)

3. If the system is within a 1-hour rainfall window with formal rainfall warnings issued → suitable=true

4. For all other scenarios, conduct comprehensive judgement to strike a balance between flood control safety and operational cost-effectiveness

Output format (JSON): {{"suitable": true/false, "reason": "supporting explanation text"}}

"""

## S3. Standardized JSON output

The JSON consists of several sections, and each section generates a set of files. The JSON outputs of AgenticAI-Control are presented as follows. All the other data are open-sourced on Github.

Github link: https://github.com/DantEzio/LLMAgent_UDSControl.

Timing_summary:

```
{
  "suitable": 0,
  "too_early": 1069,
  "total": 1069
}
```

Warning_summary:

```
{
  "total_warnings": 3,
  "total_triggers": 3,
  "warnings_file":
"C:\\Users\\chong\\Desktop\\DRL_Agent_UDSRTC\\Scenarios\\NMC_Events_2015_2025\\Event_01_2015-06-02\\Event_01_warnings.md"
}
```

v14_refresh_summary:

```
{
  "total_llm_refresh_calls": 2,
  "refresh_interval_setting": 6,
  "state_deviation_threshold": 0.15,
  "rainfall_excess_ratio": 1.5,
  "final_steps_since_last_call": 877,
  "has_baseline_state": true,
  "trigger_distribution": {
    "interval": 2,
    "deviation": 0,
    "first_event": 1,
    "unknown": 0
  }
}
```

event_info:

```
{
  "total_mm": 30.0,
  "peak_mm_hr": 7.2,
  "duration_hrs": 6,
```

"confidence": "low"
}

topo_summary:

{
Affected area: ['cc_zone'] (城关/巢湖市区排水片区)
Affected pumps: ['CC-S2', 'CC-S1', 'CC-R1', 'CC-R2']
Affected nodes: 56
}

realtime_state:

{
"node_depths": {
"HS02004478": 0.035005048462855556,
"HS02004479": 0.06872248619285606,
"HS02004480": 0.031721803718882495,
"HS02004770": 0.07791623037329216,
...
},
"link_flows": {
"HS02004478WS02006287": 9.280413138431168,
"HS02004479HS02004480": -0.007605081799820152,
"HS02004480HS02004478": 2.1274727333320866,
"HS02004770HS02004777": 0.009589123513203927,
"HS02004777HS02004785": 0.03029122097683839,
...
},
"storage_fill_status": {}
}

## S4. Case urban drainage system and rainfall data

The case study is a real-world combined urban drainage system located in eastern China, with simulation model established in the Storm Water Management Model (SWMM) encompassing 31 subcatchments, 130 pipelines, 129 manholes, and two pump stations with 7 pumps (Fig. S1) (namely the CC pump station and the JK pump station). The CC pump station has one forebay (maximum capacity 6410 $m^3$) and 4 pumps (2 pumps to downstream, 2 pumps drainage water outside the system). The JK pump station has one forebay (maximum capacity 291 $m^3$) and 3 pumps (1 pump to downstream, and 2 pumps drain water outside the system). During intense rainfall events, large volumes of stormwater enter the system, meaning that effective control must dynamically allocate flows between upstream and downstream subsystems through coordinated pump operation to reduce CSOs and flooding. In addition, water level and flow monitoring are available at a number of sites across the catchment. These include storage tanks (CC-storage, JK-storage), eleven sewer nodes (HS02004802, YS02001790, YS02001661, YS02001655, YS02001649, YS02001644, YS02001633, YS02001625, YS02001619, YS02001610, YS02001605), together with catchment-wide rainfall observations. See Fig. S1 for their locations.

This study adopts Agentic AI to acquire rainfall data and corresponding rainfall early warning information from the China Meteorological Administration (CMA, https://www.cma.gov.cn/), China Weather Network (weather.com.cn), and the open-source data platform OpenMeteo (https://open-meteo.com/). These datasets are time-aligned to construct the rainfall environment and warning signals for testing. Detailed data on rainfall intensity, warning levels and warning timings are presented in Fig. S2.

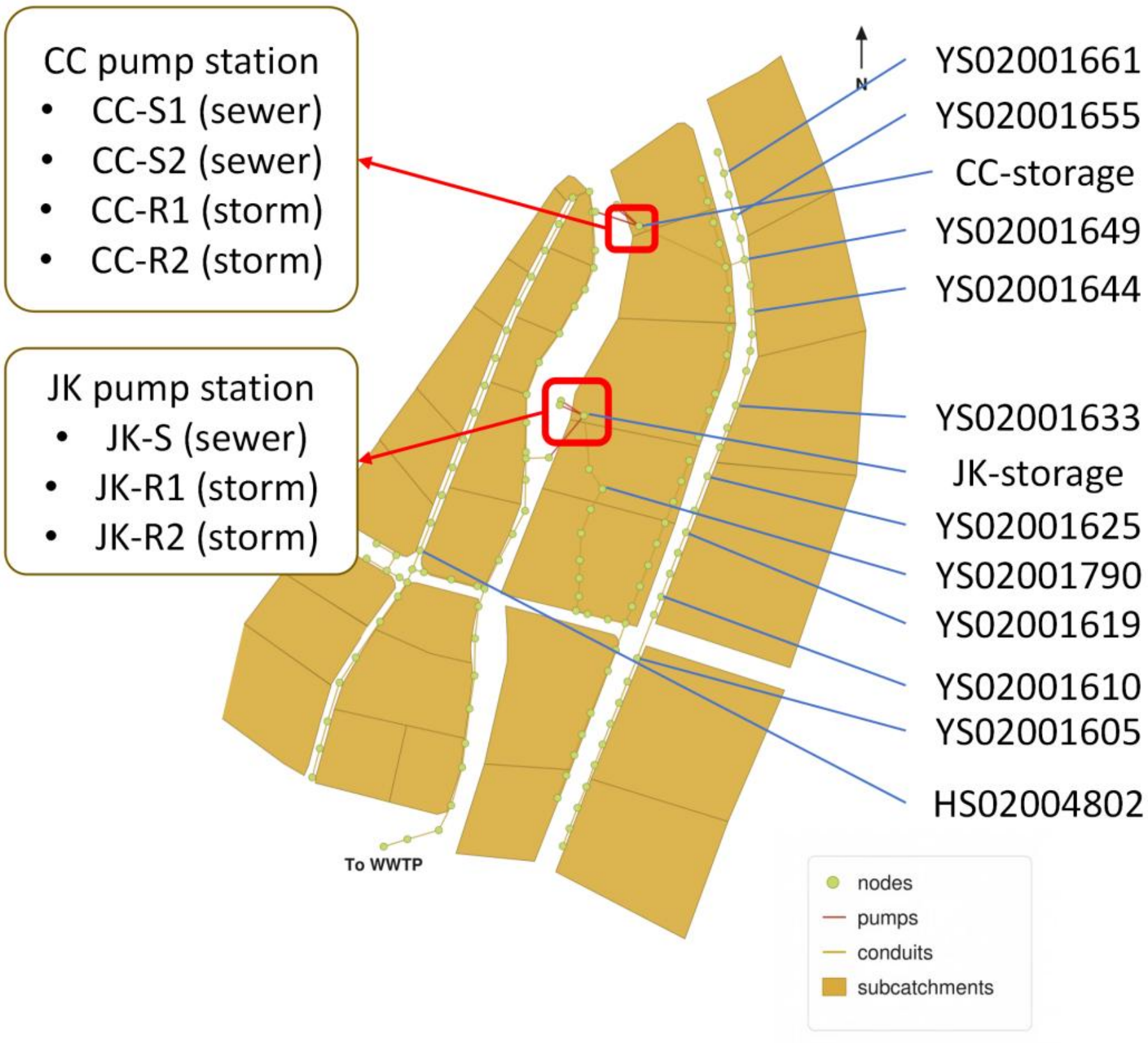

Fig. S1. SWMM of the study area.

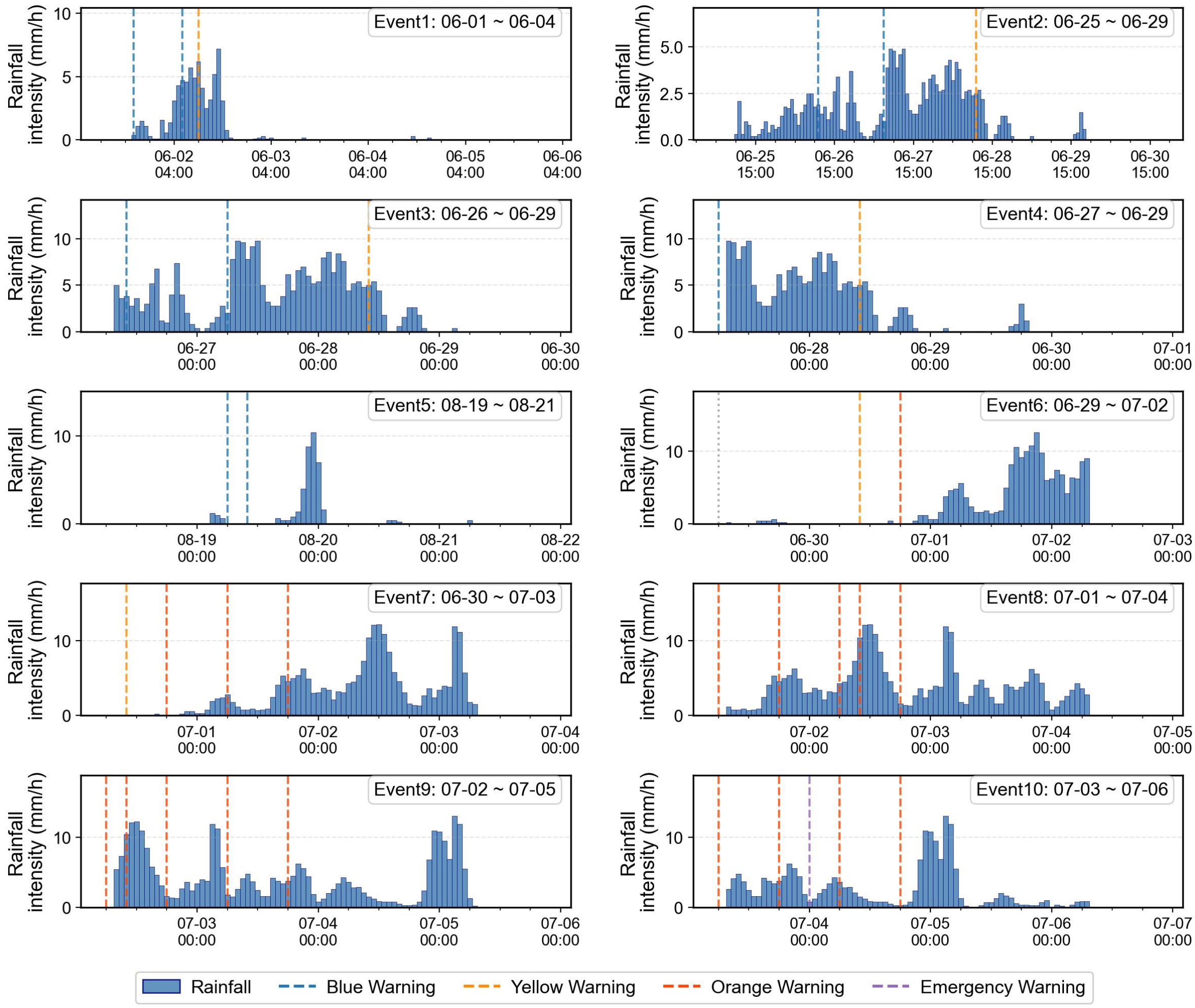

Fig. S2. Rainfall records and corresponding meteorological warning issuance timings from the official portal of the China Meteorological Administration (CMA, https://www.cma.gov.cn/), China Weather Network (weather.com.cn), and open-source data platform OpenMeteo (https://open-meteo.com/). Warnings in different colors indicate rainfall conditions of varying severity. Blue Warning: The 12-hour accumulated precipitation reaches or exceeds 50 mm, or 50 mm rainfall has occurred with ongoing precipitation. Yellow Warning: The 6-hour accumulated precipitation reaches or exceeds 50 mm, or 50 mm rainfall has occurred with ongoing precipitation. Orange Warning: The 3-hour accumulated precipitation reaches or exceeds 50 mm, or 50 mm rainfall has occurred with ongoing precipitation. Red Warning: The 3-hour accumulated precipitation reaches or exceeds 100 mm, or 100 mm rainfall has occurred with ongoing precipitation. The purple-coloured line denotes the issued emergency warning.

## S5. BC setting

BC is a rule-based control strategy which operates pumps in the study area in response to the changes of two storage tank water level. The detailed set-points are given in the Table S1 below.

Table S1. Set-points of BC.

| Pump name | Type | Pump Start water level (m) | Pump Stop water level (m) |
|---|---|---|---|
| CC-S1 | Sewage pump | 0.8 | 0.5 |
| CC-S2 | Sewage pump | 1.0 | 0.5 |
| CC-R1 | Stormwater pump | 1.2 | 0.5 |
| CC-R2 | Stormwater pump | 1.4 | 0.5 |
| JK-R1 | Stormwater pump | 4.2 | 1.2 |
| JK-R2 | Stormwater pump | 4.3 | 1.2 |
| JK-S | Sewage pump | 4.0 | 1.0 |

**S6. Meteorological early warnings details**

For the ten rainfall events selected as case studies in this research, meteorological warning data in textual form corresponding to each respective time period were collected and displayed as follows.

**Event_01 Rainfall Warning Information**

> **Event**: 2015-06-02 Blue Rainstorm Warning
> **Rainfall Time Window**: 2015-06-01 12:00 ~ 2015-06-05 12:00 (6h pre-rain + full rainfall duration + 48h post-rain)
> **Actual Rainfall Period**: 2015-06-01 18:00 ~ 2015-06-03 12:00
> **Data Source**: CMA (cma.gov.cn) Historical Warning Webpages

---

## Warnings Issued Within the Time Window

### 1. 2015-06-01 18:00 — Blue Rainstorm Warning (Continued)

| Item | Content |
|------|------|
| **Issue Time** | 2015-06-01 18:00 |
| **Issuing Agency** | China Meteorological Administration (CMA) |
| **Warning Level** | Blue Rainstorm (Level IV) |
| **Covered Areas** | Most of Hubei, western and southern Anhui, southern Jiangsu, Shanghai, northern Zhejiang, northern Jiangxi, northern Hunan, eastern and northern Guizhou |
| **Rainfall Forecast** | Localized heavy rainstorm (>100 mm) |
| **Remarks** | Continued issuance; first issued at 06:00 and maintained |

### 2. 2015-06-02 06:00 — Blue Rainstorm Warning (Continued)

| Item | Content |
|------|------|
| **Issue Time** | 2015-06-02 06:00 |
| **Issuing Agency** | China Meteorological Administration (CMA) |
| **Warning Level** | Blue Rainstorm (Level IV) |
| **Covered Areas** | Northeastern and central-southern Anhui, southern Jiangsu, Shanghai, northern Zhejiang, northeastern Jiangxi, central and northeastern Hunan, eastern and southern Guizhou, northern Guangxi |
| **Rainfall Forecast** | Localized heavy rainstorm 100–160 mm |
| **Maximum Hourly Rainfall Intensity** | 50–70 mm |
| **Remarks** | Continued issuance; rain belt persisting over the Jianghuai region |

### 3. 2015-06-02 10:00 — Yellow Rainstorm Warning (Upgraded)

| Item | Content |
|------|------|
| **Issue Time** | 2015-06-02 10:00 |
| **Issuing Agency** | China Meteorological Administration (CMA) |
| **Warning Level** | Yellow Rainstorm (Level III, upgraded from Blue) |
| **Covered Areas** | Southern Anhui, southern Jiangsu, Shanghai, northern Zhejiang, northern Jiangxi, central and northeastern Hunan, eastern and southern Guizhou, northern Guangxi |
| **Rainfall Forecast** | Localized heavy rainstorm 100–200 mm |
| **Maximum Hourly Rainfall Intensity** | 60–80 mm |
| **Remarks** | Warning level upgraded; rainfall intensity increasing |

---

## Warning Evolution Timeline
```
2015-06-01 06:00   Blue Rainstorm (First Issued)   ← Pre-window; eastern Liaoning, southeastern Henan, western Anhui, most of Hubei, etc.
2015-06-01 18:00   Blue Rainstorm (Continued)      ← ★Within window
2015-06-02 06:00   Blue Rainstorm (Continued)      ← ★Within window
2015-06-02 10:00   Yellow Rainstorm (Upgraded)     ← ★Within window
```
---
## Supplementary Local Warning Information
| Time | Agency | Warning | Content |
|------|------|------|------|
| 2015-06-01 Evening | Jiangsu Meteorological Observatory | Blue Rainstorm | Southern Jiangsu |
| 2015-06-02 Morning | Anhui Meteorological Observatory | Yellow Rainstorm | Central-southern Anhui |
---
*Data compiled: 2026-06-17 | Primary sources: CMA official website (cma.gov.cn), China Weather Network (weather.com.cn)*

**Event_02 Rainfall Warning Information**

> **Event**: 2015-06-26 Blue Rainstorm Warning
> **Rainfall Time Window**: 2015-06-25 03:00 ~ 2015-06-30 07:00 (6h pre-rain + full rainfall duration + 48h post-rain)
> **Actual Rainfall Period**: 2015-06-25 09:00 ~ 2015-06-28 07:00
> **Data Source**: CMA (cma.gov.cn) Historical Warning Webpages, Xinhua News
---
## Warnings Issued Within the Time Window
### 1. 2015-06-26 10:00 — Blue Rainstorm Warning (First Issued)
| Item | Content |
|------|------|
| **Issue Time** | 2015-06-26 10:00 |
| **Issuing Agency** | China Meteorological Administration (CMA) |
| **Warning Level** | Blue Rainstorm (Level IV) |
| **Covered Areas** | Northwestern and southeastern Henan, **central Anhui**, **southern Jiangsu**, Shanghai, northern Zhejiang, northern Sichuan Basin, central Shaanxi, southwestern Shanxi |
| **Rainfall Forecast** | Localized heavy rainstorm 100–130 mm |
| **Maximum Hourly Rainfall Intensity** | 50–70 mm |
| **Remarks** | Rain belt shifting northward to the Huai River basin; Jiangsu and Anhui becoming the most concentrated heavy rainfall areas |
### 2. 2015-06-27 06:00 — Blue Rainstorm Warning (Continued)
| Item | Content |
|------|------|
| **Issue Time** | 2015-06-27 06:00 |
| **Issuing Agency** | China Meteorological Administration (CMA) |
| **Warning Level** | Blue Rainstorm (Level IV) |
| **Covered Areas** | Southeastern Henan, northeastern Hubei, **most of Anhui**, **central-southern Jiangsu**, Shanghai, northern Zhejiang, southwestern Shaanxi, northern Sichuan Basin |
| **Rainfall Forecast** | Heavy rainstorm 100–120 mm in southeastern Henan, central and southeastern Anhui, southern

Jiangsu, etc. |
| **Maximum Hourly Rainfall Intensity** | 40–60 mm |
| **Remarks** | Rainfall intensity continuing to intensify |
### 3. 2015-06-28 10:00 — Yellow Rainstorm Warning (Upgraded)
| Item | Content |
|------|------|
| **Issue Time** | 2015-06-28 10:00 |
| **Issuing Agency** | China Meteorological Administration (CMA) |
| **Warning Level** | Yellow Rainstorm (Level III, upgraded from Blue) |
| **Covered Areas** | Southeastern Henan, **northern and eastern Anhui**, **southern and western Jiangsu**, Shanghai, northeastern Zhejiang, southwestern Shaanxi, central-northern Sichuan Basin |
| **Rainfall Forecast** | Localized heavy rainstorm 100–180 mm in northern Sichuan Basin and southwestern Shaanxi |
| **Maximum Hourly Rainfall Intensity** | 50–70 mm |
| **Precondition** | In the past 36 hours, 150–250 mm of heavy rainfall had already occurred in southern Jiangsu, central Anhui, southeastern Henan, etc., with localized accumulations exceeding 300 mm |
| **Remarks** | Anhui rainstorm warning upgraded from Blue to Yellow |
---
## Warning Evolution Timeline
```
2015-06-26 10:00   Blue Rainstorm (First Issued)    ← ★Within window
2015-06-27 06:00   Blue Rainstorm (Continued)       ← ★Within window
2015-06-28 10:00   Yellow Rainstorm (Upgraded)      ← ★Within window
```
---
## Local Warning Information
| Time | Agency | Warning | Content |
|------|------|------|------|
| 2015-06-26 | Anhui Meteorological Bureau | Major Meteorological Disaster (Rainstorm) Level IV Emergency Response | Activated; focus on strengthening rainfall monitoring along the Huai River main stem |
| 2015-06-26 19:33 | Jiangsu Meteorological Observatory | Blue Rainstorm Warning | Province-wide |
| 2015-06-27 03:34 | Jiangsu Meteorological Observatory | Orange Rainstorm Warning | Upgraded; Nanjing, Changzhou, Yangzhou, etc. |
---
*Data compiled: 2026-06-17 | Primary sources: CMA official website (cma.gov.cn), Xinhua News, China Weather Network*

**Event_03 Rainfall Warning Information**
> **Event**: 2015-06-27 Blue Rainstorm Warning
> **Rainfall Time Window**: 2015-06-26 02:00 ~ 2015-07-01 03:00 (6h pre-rain + full rainfall duration + 48h post-rain)
> **Actual Rainfall Period**: 2015-06-26 08:00 ~ 2015-06-29 03:00
> **Data Source**: CMA (cma.gov.cn) Historical Warning Webpages
> **Note**: This event, together with Event_02 (2015-06-26) and Event_04 (2015-06-28), belongs to the same round of heavy rainfall; warning information is highly overlapping.
---
## Warnings Issued Within the Time Window
### 1. 2015-06-26 10:00 — Blue Rainstorm Warning (First Issued)

| Item | Content |
|------|------|
| **Issue Time** | 2015-06-26 10:00 |
| **Issuing Agency** | China Meteorological Administration (CMA) |
| **Warning Level** | Blue Rainstorm (Level IV) |
| **Covered Areas** | Northwestern and southeastern Henan, **central Anhui**, **southern Jiangsu**, Shanghai, northern Zhejiang |
| **Rainfall Forecast** | Localized heavy rainstorm 100–130 mm |
| **Remarks** | First warning of this round of heavy rainfall |

### 2. 2015-06-27 06:00 — Blue Rainstorm Warning (Continued)

| Item | Content |
|------|------|
| **Issue Time** | 2015-06-27 06:00 |
| **Issuing Agency** | China Meteorological Administration (CMA) |
| **Warning Level** | Blue Rainstorm (Level IV) |
| **Covered Areas** | Southeastern Henan, northeastern Hubei, **most of Anhui**, **central-southern Jiangsu**, Shanghai, northern Zhejiang |
| **Rainfall Forecast** | Heavy rainstorm 100–120 mm in central and southeastern Anhui, southern Jiangsu, etc. |
| **Remarks** | Core warning day of this event |

### 3. 2015-06-28 10:00 — Yellow Rainstorm Warning (Upgraded)

| Item | Content |
|------|------|
| **Issue Time** | 2015-06-28 10:00 |
| **Issuing Agency** | China Meteorological Administration (CMA) |
| **Warning Level** | Yellow Rainstorm (Level III, upgraded from Blue) |
| **Covered Areas** | Southeastern Henan, **northern and eastern Anhui**, **southern and western Jiangsu**, Shanghai, northeastern Zhejiang |
| **Rainfall Forecast** | Localized heavy rainstorm 100–180 mm |
| **Precondition** | In the past 36 hours, 150–250 mm of heavy rainfall had already occurred in southern Jiangsu, central Anhui, etc. |

---

## Warning Evolution Timeline

```
2015-06-26 10:00   Blue Rainstorm (First Issued)    ← ★Within window
2015-06-27 06:00   Blue Rainstorm (Continued)       ← ★Within window (Core warning day)
2015-06-28 10:00   Yellow Rainstorm (Upgraded)      ← ★Within window
```

---

## Local Warning Information

| Time | Agency | Warning | Content |
|------|------|------|------|
| 2015-06-26 | Anhui Meteorological Bureau | Level IV Emergency Response | Activated Major Meteorological Disaster (Rainstorm) emergency response |
| 2015-06-26 19:33 | Jiangsu Meteorological Observatory | Blue Rainstorm | Province-wide |
| 2015-06-27 03:34 | Jiangsu Meteorological Observatory | Orange Rainstorm | Upgraded; Nanjing, Changzhou, Yangzhou, etc. issued Red Rainstorm warnings |

| 2015-06-27 09:30 | Jiangsu Meteorological Observatory | Orange Rainstorm (Maintained) | Persistent heavy rainfall in southern Jiangsu |
---
## Related Events
This event is part of the June 26–29, 2015 Huai River basin heavy rainfall process. Related events:
- Event_02 (2015-06-26): First day of the same process
- Event_04 (2015-06-28): Subsequent upgrade of the same process
---
*Data compiled: 2026-06-17 | Primary sources: CMA official website (cma.gov.cn), Xinhua News, Wikiwand*

**Event_04 Rainfall Warning Information**
> **Event**: 2015-06-28 Blue Rainstorm Warning
> **Rainfall Time Window**: 2015-06-27 02:00 ~ 2015-07-01 19:00 (6h pre-rain + full rainfall duration + 48h post-rain)
> **Actual Rainfall Period**: 2015-06-27 08:00 ~ 2015-06-29 19:00
> **Data Source**: CMA (cma.gov.cn) Historical Warning Webpages
> **Note**: This event, together with Event_02 (2015-06-26) and Event_03 (2015-06-27), belongs to the same round of heavy rainfall; warning information is highly overlapping.
---
## Warnings Issued Within the Time Window
### 1. 2015-06-27 06:00 — Blue Rainstorm Warning (Continued)
| Item | Content |
|------|------|
| **Issue Time** | 2015-06-27 06:00 |
| **Issuing Agency** | China Meteorological Administration (CMA) |
| **Warning Level** | Blue Rainstorm (Level IV) |
| **Covered Areas** | Southeastern Henan, northeastern Hubei, **most of Anhui**, **central-southern Jiangsu**, Shanghai, northern Zhejiang, southwestern Shaanxi, northern Sichuan Basin |
| **Rainfall Forecast** | Heavy rainstorm 100–120 mm in southeastern Henan, central and southeastern Anhui, southern Jiangsu, etc. |
| **Maximum Hourly Rainfall Intensity** | 40–60 mm |
### 2. 2015-06-28 10:00 — Yellow Rainstorm Warning (Upgraded)
| Item | Content |
|------|------|
| **Issue Time** | 2015-06-28 10:00 |
| **Issuing Agency** | China Meteorological Administration (CMA) |
| **Warning Level** | Yellow Rainstorm (Level III) |
| **Covered Areas** | Southeastern Henan, **northern and eastern Anhui**, **southern and western Jiangsu**, Shanghai, northeastern Zhejiang, southwestern Shaanxi, central-northern Sichuan Basin |
| **Rainfall Forecast** | Localized heavy rainstorm 100–180 mm in northern Sichuan Basin and southwestern Shaanxi |
| **Precondition** | In the past 36 hours, 150–250 mm of heavy rainfall had already occurred in southern Jiangsu, central Anhui, southeastern Henan, etc., with localized accumulations exceeding 300 mm |
| **Remarks** | Core warning day of this event; warning upgraded from Blue to Yellow |
---
## Warning Evolution Timeline
```
2015-06-27 06:00    Blue Rainstorm (Continued)    ← ★Within window
```

2015-06-28 10:00    Yellow Rainstorm (Upgraded) ← ★Within window (Core warning day)
```
---
## Local Warning Information
| Time | Agency | Warning | Content |
|------|------|------|------|
| 2015-06-28 | Jiangsu Meteorological Observatory | Orange Rainstorm (Maintained) | Southern and central Jiangsu |
| 2015-06-28 | Anhui Meteorological Observatory | Yellow Rainstorm | Northern and eastern Anhui |
---
## Related Events
This event is the concluding phase of the June 26–29, 2015 Huai River basin heavy rainfall process. Related events:
- Event_02 (2015-06-26): First day of the same process
- Event_03 (2015-06-27): Middle phase of the same process
---
*Data compiled: 2026-06-17 | Primary sources: CMA official website (cma.gov.cn)*

**Event_05 Rainfall Warning Information**
> **Event**: 2015-08-19 Blue Rainstorm Warning
> **Rainfall Time Window**: 2015-08-18 21:00 ~ 2015-08-23 06:00 (6h pre-rain + full rainfall duration + 48h post-rain)
> **Actual Rainfall Period**: 2015-08-19 03:00 ~ 2015-08-21 06:00
> **Data Source**: CMA (cma.gov.cn), China Weather Network (weather.com.cn), Sina News
---
## Pre-Window Warnings (For Reference)
### 2015-08-18 06:00 — Blue Rainstorm Warning
| Item | Content |
|------|------|
| **Issue Time** | 2015-08-18 06:00 |
| **Issuing Agency** | China Meteorological Administration (CMA) |
| **Warning Level** | Blue Rainstorm (Level IV) |
| **Covered Areas** | Northeastern Sichuan Basin, northern and eastern Chongqing, most of Hubei, southern Henan, **central Anhui**, northern Hunan, southern Jiangsu |
| **Remarks** | Pre-window warning; southwest vortex moving eastward triggering this round of heavy rainfall |
### 2015-08-18 18:00 — Blue Rainstorm Warning (Continued)
| Item | Content |
|------|------|
| **Issue Time** | 2015-08-18 18:00 |
| **Issuing Agency** | China Meteorological Administration (CMA) |
| **Warning Level** | Blue Rainstorm (Level IV) |
| **Covered Areas** | Southeastern Chongqing, most of Hubei, northern Hunan, southern Henan, **central-northern Anhui**, central-southern Jiangsu |
| **Rainfall Forecast** | Localized heavy rainstorm 100–140 mm in northeastern Hubei and northwestern Hunan |
---
## Warnings Issued Within the Time Window
### 1. 2015-08-19 06:00 — Blue Rainstorm Warning (Continued)
| Item | Content |
|------|------|
```

| **Issue Time** | 2015-08-19 06:00 |
| **Issuing Agency** | China Meteorological Administration (CMA) |
| **Warning Level** | Blue Rainstorm (Level IV) |
| **Covered Areas** | Northeastern Hubei, southeastern Henan, northeastern Hunan, northwestern Jiangxi, **central-northern Anhui**, **western Jiangsu**, southeastern and northwestern Yunnan, northwestern Guangxi |
| **Rainfall Forecast** | Localized heavy rainstorm 100–120 mm in eastern Anhui and western Jiangsu |
| **Maximum Hourly Rainfall Intensity** | 40–60 mm |
| **Remarks** | ★First warning within window |

### 2. 2015-08-19 10:00 — Blue Rainstorm Warning (Continued)

| Item | Content |
|------|------|
| **Issue Time** | 2015-08-19 10:00 |
| **Issuing Agency** | China Meteorological Administration (CMA) |
| **Warning Level** | Blue Rainstorm (Level IV) |
| **Covered Areas** | **Central and southern Anhui**, **central-southern Jiangsu**, northern and eastern Jiangxi, northwestern Fujian, central Guangdong, western Guangxi, central-western and southern Yunnan |
| **Rainfall Forecast** | Localized heavy rainstorm 100–180 mm in east-central Anhui and central Jiangsu |
| **Maximum Hourly Rainfall Intensity** | 40–70 mm |
| **Remarks** | Rainfall intensity forecast increased compared to 06:00; east-central Anhui to central Jiangsu as the rainstorm center |

---

## Warning Evolution Timeline

```
2015-08-18 06:00   Blue Rainstorm (First Issued)   ← Pre-window
2015-08-18 18:00   Blue Rainstorm (Continued)      ← Pre-window
2015-08-19 06:00   Blue Rainstorm (Continued)      ← ★Within window
2015-08-19 10:00   Blue Rainstorm (Continued)      ← ★Within window; rainstorm center locked on Anhui-Jiangsu
```

---

## Local Warning Information

| Time | Agency | Warning | Content |
|------|------|------|------|
| 2015-08-19 | Anhui Meteorological Observatory | Blue/Yellow Rainstorm | Central Anhui |

---

*Data compiled: 2026-06-17 | Primary sources: CMA official website, China Weather Network (weather.com.cn), Sina News*

**Event_06 Rainfall Warning Information**

> **Event**: 2016-06-30 Orange Rainstorm Warning — First of 2016
> **Rainfall Time Window**: 2016-06-29 02:00 ~ 2016-07-04 07:00 (6h pre-rain + full rainfall duration + 48h post-rain)
> **Actual Rainfall Period**: 2016-06-29 08:00 ~ 2016-07-02 07:00
> **Data Source**: CMA (cma.gov.cn), Xinhua News, China.com
> **Historical Significance**: This was the **first Orange Rainstorm Warning** issued by the CMA since the start of the 2016 flood season, marking the onset of the most intense rainfall phase in the middle and lower reaches of the Yangtze River.

---

## Pre-Window Warnings (For Reference)
### 2016-06-28 10:00 — Yellow Rainstorm Warning (Continued)
| Item | Content |
|------|------|
| **Issue Time** | 2016-06-28 10:00 |
| **Issuing Agency** | China Meteorological Administration (CMA) |
| **Warning Level** | Yellow Rainstorm (Level III) |
| **Covered Areas** | Northern Guizhou, northern Hunan, eastern Hubei, **southern Anhui**, northern and southwestern Jiangxi, **southern Jiangsu**, western and northern Zhejiang, Shanghai |
| **Rainfall Forecast** | Localized heavy rainstorm 100–150 mm in northwestern Hunan, southeastern Anhui, southern Jiangsu, and northwestern Zhejiang |
| **Maximum Hourly Rainfall Intensity** | 40–70 mm |

---
## Warnings Issued Within the Time Window
### 1. 2016-06-29 06:00 — Yellow Rainstorm Warning Lifted
| Item | Content |
|------|------|
| **Issue Time** | 2016-06-29 06:00 |
| **Issuing Agency** | China Meteorological Administration (CMA) |
| **Warning Level** | Yellow Rainstorm Warning Lifted |
| **Reason** | Rainstorm coverage significantly reduced, intensity weakened |
| **Remarks** | Scattered heavy rain still expected from 08:00 on the 29th to 08:00 on the 30th, but no longer meeting warning criteria. This was a brief intermission. |

### 2. 2016-06-30 10:00 — Yellow Rainstorm Warning (Re-issued)
| Item | Content |
|------|------|
| **Issue Time** | 2016-06-30 10:00 |
| **Issuing Agency** | China Meteorological Administration (CMA) |
| **Warning Level** | Yellow Rainstorm (Level III) |
| **Covered Areas** | Eastern Southwest China, most of Jianghan, most of Jianghuai, southern and eastern Huanghuai, and northwestern Jiangnan |
| **Core Areas** | Central Chongqing, northwestern Hunan, southwestern and eastern Hubei, southeastern Henan, **central-northern Anhui**, **central-northern Jiangsu** |
| **Rainfall Forecast** | Heavy rainstorm 100–180 mm in central-northern Anhui, central-northern Jiangsu, etc. |
| **Maximum Hourly Rainfall Intensity** | 60–90 mm |
| **Accompanying Warning** | Concurrently issued Blue Severe Convective Weather Warning: most of Anhui and Jiangsu with short-duration heavy rainfall |

### 3. 2016-06-30 18:00 — Orange Rainstorm Warning (Upgraded, First of the Year!)
| Item | Content |
|------|------|
| **Issue Time** | 2016-06-30 18:00 |
| **Issuing Agency** | China Meteorological Administration (CMA) |
| **Warning Level** | **Orange Rainstorm (Level II)** — First of 2016! |
| **Covered Areas** | Most of Hubei, **central-northern Anhui**, **central-northern Jiangsu**, southern Henan |
| **Rainfall Forecast** | Heavy rainstorm 100–200 mm in northeastern and southwestern Hubei, central Anhui, central Jiangsu, and southern Henan |

| **Extreme Forecast** | **Localized extraordinary rainstorm 250–280 mm** |
| **Maximum Hourly Rainfall Intensity** | 60–90 mm |
| **Historical Status** | This was the **first Orange Rainstorm Warning** issued by the CMA since the start of the 2016 flood season, marking the beginning of the most intense rainfall phase in the middle and lower Yangtze |
---
## Warning Evolution Timeline
```
2016-06-28 10:00    Yellow Rainstorm (Continued)                       ← Pre-window
2016-06-29 06:00    Yellow Rainstorm Lifted                            ← ★Within window (Brief intermission)
2016-06-30 10:00    Yellow Rainstorm (Re-issued)                       ← ★Within window
2016-06-30 18:00    Orange Rainstorm (Upgraded, First of Year)  ← ★Within window ★Core event
```
---
## Characteristics of This Round
- **Leapfrog Warning Upgrade**: Within less than 8 hours, the warning escalated directly from Yellow to Orange (first of the year), reflecting a sharp increase in rainfall intensity
- **Outbreak After Brief Intermission**: Only 1 day after the warning was lifted on June 29, it was re-issued and rapidly upgraded on the 30th as the Meiyu front intensified again
- **Severe Convection Accompaniment**: Concurrent Blue Severe Convective Weather Warning issued, with hourly rainfall intensity reaching 60–90 mm
---
*Data compiled: 2026-06-17 | Primary sources: CMA official website (cma.gov.cn), Xinhua News, China.com, China Weather Network*

**Event_07 Rainfall Warning Information**
> **Event**: 2016-07-01 Orange Rainstorm Warning (Continued)
> **Rainfall Time Window**: 2016-06-30 10:00 ~ 2016-07-05 07:00 (6h pre-rain + full rainfall duration + 48h post-rain)
> **Actual Rainfall Period**: 2016-06-30 16:00 ~ 2016-07-03 07:00
> **Data Source**: CMA (cma.gov.cn), China Weather Network Weibo, CCTV
> **Note**: This event, together with Event_06 (2016-06-30) and Events 08–10, is part of the June 30 – July 6, 2016 historically severe rainfall process in the middle and lower Yangtze River.
---
## Warnings Issued Within the Time Window
### 1. 2016-06-30 10:00 — Yellow Rainstorm Warning
| Item | Content |
|------|------|
| **Issue Time** | 2016-06-30 10:00 |
| **Issuing Agency** | China Meteorological Administration (CMA) |
| **Warning Level** | Yellow Rainstorm (Level III) |
| **Covered Areas** | **Central-northern Anhui**, **central-northern Jiangsu**, southwestern and eastern Hubei, southeastern Henan, central Chongqing, northwestern Hunan |
| **Rainfall Forecast** | Heavy rainstorm 100–180 mm in central-northern Anhui, central-northern Jiangsu, etc. |
| **Maximum Hourly Rainfall Intensity** | 60–90 mm |
### 2. 2016-06-30 18:00 — Orange Rainstorm Warning (Upgraded)
| Item | Content |
|------|------|

| **Issue Time** | 2016-06-30 18:00 |
| **Issuing Agency** | China Meteorological Administration (CMA) |
| **Warning Level** | Orange Rainstorm (Level II, upgraded from Yellow) |
| **Covered Areas** | Most of Hubei, **central-northern Anhui**, **central-northern Jiangsu**, southern Henan |
| **Rainfall Forecast** | Heavy rainstorm 100–200 mm in northeastern and southwestern Hubei, central Anhui, and central Jiangsu |
| **Extreme Forecast** | Localized extraordinary rainstorm 250–280 mm |

### 3. 2016-07-01 06:00 — Orange Rainstorm Warning (Continued)

| Item | Content |
|------|------|
| **Issue Time** | 2016-07-01 06:00 |
| **Issuing Agency** | China Meteorological Administration (CMA) |
| **Warning Level** | Orange Rainstorm (Level II) |
| **Covered Areas** | **Most of Jiangsu and Anhui**, central-eastern Hubei, central-northern Hunan, northern Jiangxi, southern Guizhou, northern Guangxi, western Yunnan |
| **Core Areas** | Southwestern Anhui, eastern Hubei, western and northeastern Hunan, northern Jiangxi, southeastern Guizhou |
| **Rainfall Forecast** | Heavy rainstorm magnitude |
| **Remarks** | Orange Warning entering its second day |

### 4. 2016-07-01 18:00 — Orange Rainstorm Warning (Continued)

| Item | Content |
|------|------|
| **Issue Time** | 2016-07-01 18:00 |
| **Issuing Agency** | China Meteorological Administration (CMA) |
| **Warning Level** | Orange Rainstorm (Level II) |
| **Covered Areas** | **Central Anhui**, **central-southern Jiangsu**, eastern Hubei, northeastern and southwestern Hunan, southeastern Guizhou |
| **Rainfall Forecast** | Heavy rainstorm 100–180 mm in central Anhui, central Jiangsu, northeastern Hunan, and southeastern Guizhou |
| **Maximum Hourly Rainfall Intensity** | 60–80 mm |

---

## Warning Evolution Timeline

```
2016-06-30 10:00    Yellow Rainstorm                ← ★Within window
2016-06-30 18:00    Orange Rainstorm (Upgraded)      ← ★Within window, First Orange of the year
2016-07-01 06:00    Orange Rainstorm (Continued)    ← ★Within window
2016-07-01 18:00    Orange Rainstorm (Continued)    ← ★Within window
```

---

## Related Events

This event is part of the June 30 – July 6, 2016 historically severe rainfall process in the middle and lower Yangtze River. Related events:

- Event_06 (2016-06-30): First day of Orange Warning
- Event_08 (2016-07-02): Day 3 of Orange Warning
- Event_09 (2016-07-03): Day 4 of Orange Warning
- Event_10 (2016-07-04): Major Meteorological Disaster Level II Emergency Response

---
*Data compiled: 2026-06-17 | Primary sources: CMA official website (cma.gov.cn), China Weather Network Weibo, CCTV*

**Event_08 Rainfall Warning Information**

> **Event**: 2016-07-02 Orange Rainstorm Warning (3rd Consecutive Day)
> **Rainfall Time Window**: 2016-07-01 02:00 ~ 2016-07-06 07:00 (6h pre-rain + full rainfall duration + 48h post-rain)
> **Actual Rainfall Period**: 2016-07-01 08:00 ~ 2016-07-04 07:00
> **Data Source**: CMA (cma.gov.cn), Sohu News, CCTV

---
## Warnings Issued Within the Time Window
### 1. 2016-07-01 06:00 — Orange Rainstorm Warning (Continued)
| Item | Content |
|------|------|
| **Issue Time** | 2016-07-01 06:00 |
| **Issuing Agency** | China Meteorological Administration (CMA) |
| **Warning Level** | Orange Rainstorm (Level II) |
| **Covered Areas** | **Most of Jiangsu and Anhui**, central-eastern Hubei, central-northern Hunan, northern Jiangxi, southern Guizhou, northern Guangxi |
| **Remarks** | Orange Warning Day 2 |

### 2. 2016-07-01 18:00 — Orange Rainstorm Warning (Continued)
| Item | Content |
|------|------|
| **Issue Time** | 2016-07-01 18:00 |
| **Issuing Agency** | China Meteorological Administration (CMA) |
| **Warning Level** | Orange Rainstorm (Level II) |
| **Covered Areas** | **Central Anhui**, **central-southern Jiangsu**, eastern Hubei, northeastern and southwestern Hunan, southeastern Guizhou |
| **Rainfall Forecast** | Heavy rainstorm 100–180 mm in central Anhui and central Jiangsu |

### 3. 2016-07-02 06:00 — Orange Rainstorm Warning (3rd Consecutive Day) ★Core
| Item | Content |
|------|------|
| **Issue Time** | 2016-07-02 06:00 |
| **Issuing Agency** | China Meteorological Administration (CMA) |
| **Warning Level** | Orange Rainstorm (Level II, 3rd consecutive day) |
| **Covered Areas** | **Central-southern Anhui**, **central-southern Jiangsu**, eastern Hubei, northern Hunan, northern Jiangxi, southeastern Guizhou, northern Guangxi |
| **Rainfall Forecast** | Heavy rainstorm 100–240 mm in central Anhui, central Jiangsu, northern Hunan, etc. |
| **Precondition** | Since June 30, Jianghan and Jianghuai regions have seen relentless rainfall; the heavy rainstorm area across Hubei, Anhui, and Jiangsu has reached 140,000 km$^2$ |
| **Remarks** | Core warning day of this event; Orange Warning entering Day 3; localized accumulations in eastern Hubei and central Anhui exceeding 400 mm |

### 4. 2016-07-02 10:00 — Orange Rainstorm Warning (Continued)
| Item | Content |
|------|------|
| **Issue Time** | 2016-07-02 10:00 |

| **Issuing Agency** | China Meteorological Administration (CMA) |
| **Warning Level** | Orange Rainstorm (Level II) |
| **Covered Areas** | **Central-southern Anhui**, **central-southern Jiangsu**, eastern Hubei, northern Hunan, northern Jiangxi, southeastern Guizhou, northern Guangxi |
| **Rainfall Forecast** | Localized heavy rainstorm 100–240 mm |

### 5. 2016-07-02 18:00 — Orange Rainstorm Warning (Continued)

| Item | Content |
|------|------|
| **Issue Time** | 2016-07-02 18:00 |
| **Issuing Agency** | China Meteorological Administration (CMA) |
| **Warning Level** | Orange Rainstorm (Level II) |
| **Covered Areas** | **Central-southern Anhui**, **central-southern Jiangsu**, eastern Hubei, etc. |

---

## Warning Evolution Timeline

```
2016-07-01 06:00   Orange Rainstorm (Continued)   ← ★Within window (Orange Day 2)
2016-07-01 18:00   Orange Rainstorm (Continued)   ← ★Within window
2016-07-02 06:00   Orange Rainstorm (Day 3)       ← ★Within window ★Core
2016-07-02 10:00   Orange Rainstorm (Continued)   ← ★Within window
2016-07-02 18:00   Orange Rainstorm (Continued)   ← ★Within window
```

---

## Disaster Context

As of July 2:

- Localized cumulative rainfall in eastern Hubei and central Anhui exceeded 400 mm
- Heavy rainstorm area across Hubei, Anhui, and Jiangsu reached 140,000 $km^2$
- Mountain torrents broke out in Jinzhai, Anhui; Hanshan activated Level II response
- Tongcheng, Anhui experienced extraordinary rainstorm (June 28 – July 2), with maximum point rainfall of 397.7 mm

---

## Related Events

Part of the June 30 – July 6, 2016 historically severe rainfall process in the middle and lower Yangtze River.

---

*Data compiled: 2026-06-17 | Primary sources: CMA official website (cma.gov.cn), Sohu News, CCTV*

**Event_09 Rainfall Warning Information**

> **Event**: 2016-07-03 Orange Rainstorm Warning (Continued)
> **Rainfall Time Window**: 2016-07-02 02:00 ~ 2016-07-07 07:00 (6h pre-rain + full rainfall duration + 48h post-rain)
> **Actual Rainfall Period**: 2016-07-02 08:00 ~ 2016-07-05 07:00
> **Data Source**: CMA (cma.gov.cn), Xinhua News, Jiemian News

---

## Warnings Issued Within the Time Window

### 1. 2016-07-02 06:00 — Orange Rainstorm Warning (3rd Consecutive Day)

| Item | Content |
|------|------|
| **Issue Time** | 2016-07-02 06:00 |
| **Issuing Agency** | China Meteorological Administration (CMA) |

| **Warning Level** | Orange Rainstorm (Level II) |
| **Covered Areas** | **Central-southern Anhui**, **central-southern Jiangsu**, eastern Hubei, northern Hunan, northern Jiangxi, southeastern Guizhou, northern Guangxi |
| **Rainfall Forecast** | Heavy rainstorm 100–240 mm in central Anhui, central Jiangsu, and northern Hunan |

### 2. 2016-07-02 10:00 — Orange Rainstorm Warning (Continued)

| Item | Content |
|------|------|
| **Issue Time** | 2016-07-02 10:00 |
| **Warning Level** | Orange Rainstorm (Level II) |
| **Covered Areas** | **Central-southern Anhui**, **central-southern Jiangsu**, eastern Hubei, northern Hunan, etc. |
| **Rainfall Forecast** | Localized heavy rainstorm 100–240 mm |

### 3. 2016-07-02 18:00 — Orange Rainstorm Warning (Continued)

| Item | Content |
|------|------|
| **Issue Time** | 2016-07-02 18:00 |
| **Warning Level** | Orange Rainstorm (Level II) |

### 4. 2016-07-03 06:00 — Orange Rainstorm Warning (Continued) ★Core

| Item | Content |
|------|------|
| **Issue Time** | 2016-07-03 06:00 |
| **Issuing Agency** | China Meteorological Administration (CMA) |
| **Warning Level** | Orange Rainstorm (Level II) |
| **Covered Areas** | **Central-southern Anhui**, **central-southern Jiangsu**, eastern Hubei, northwestern Jiangxi, central-southern Hunan, southeastern Guizhou, northern Guangxi |
| **Rainfall Forecast** | Heavy rainstorm 100–220 mm along the Yangtze in Anhui and Jiangsu, southeastern Hubei, northwestern Jiangxi, central Hunan, and eastern Guizhou |
| **Maximum Hourly Rainfall Intensity** | 50–80 mm |
| **Cumulative Rainfall** | 100–260 mm in eastern Hubei, along the Yangtze in Jiangsu and Anhui, and northeastern Hunan, locally exceeding 300 mm |
| **Precondition** | From June 30 to 05:00 July 3, 100–300 mm of heavy rainfall had already occurred in central-eastern Hubei, central-southern Anhui, and central-southern Jiangsu, with localized areas in eastern Hubei and central Anhui exceeding 400 mm |
| **Remarks** | Core warning day of this event. Strong rainfall will continue in the above areas over the next two days; watch for flooding and landslides caused by persistent heavy rainfall |

### 5. 2016-07-03 18:00 — Orange Rainstorm Warning (Continued)

| Item | Content |
|------|------|
| **Issue Time** | 2016-07-03 18:00 |
| **Issuing Agency** | China Meteorological Administration (CMA) |
| **Warning Level** | Orange Rainstorm (Level II) |
| **Covered Areas** | **Central-southern Anhui**, **central-southern Jiangsu**, etc. |
| **Remarks** | Day 4 of Orange Warning |

---

## Warning Evolution Timeline

```
2016-07-02 06:00    Orange Rainstorm (Day 3)       ← ★Within window
```

2016-07-02 10:00　Orange Rainstorm (Continued) ← ★Within window
2016-07-02 18:00　Orange Rainstorm (Continued) ← ★Within window
2016-07-03 06:00　Orange Rainstorm (Continued) ← ★Within window ★Core (Heavy rainstorm 100–220 mm)
2016-07-03 18:00　Orange Rainstorm (Continued) ← ★Within window (Day 4)
```
---
## Future Trends
From July 5 to 8, the main rain belt will shift somewhat northward; heavy to rainstorm-level rainfall is expected in eastern Southwest China, Jianghan, central-southern Huanghuai, and areas along the Huai River, with localized heavy rainstorms in some areas.
---
## Related Events
Part of the June 30 – July 6, 2016 historically severe rainfall process in the middle and lower Yangtze River.
---
*Data compiled: 2026-06-17 | Primary sources: CMA official website (cma.gov.cn), Xinhua News, Jiemian News*

**Event_10 Rainfall Warning Information**
> **Event**: 2016-07-04 Major Meteorological Disaster Level II Emergency Response
> **Rainfall Time Window**: 2016-07-03 02:00 ~ 2016-07-08 07:00 (6h pre-rain + full rainfall duration + 48h post-rain)
> **Actual Rainfall Period**: 2016-07-03 08:00 ~ 2016-07-06 07:00
> **Data Source**: CMA (cma.gov.cn), Xinhua News, China Meteorological Administration
> **Historical Significance**: Due to persistent heavy rainfall in the middle and lower Yangtze River, the China Meteorological Administration activated the **Major Meteorological Disaster (Rainstorm) Level II Emergency Response**, the highest-level response measure of this round.
---
## Warnings Issued Within the Time Window
### 1. 2016-07-03 06:00 — Orange Rainstorm Warning (Continued)
| Item | Content |
|------|------|
| **Issue Time** | 2016-07-03 06:00 |
| **Issuing Agency** | China Meteorological Administration (CMA) |
| **Warning Level** | Orange Rainstorm (Level II) |
| **Covered Areas** | **Central-southern Anhui**, **central-southern Jiangsu**, eastern Hubei, northwestern Jiangxi, central-southern Hunan, southeastern Guizhou, northern Guangxi |
| **Rainfall Forecast** | Heavy rainstorm 100–220 mm along the Yangtze in Anhui and Jiangsu |
| **Cumulative Rainfall** | 100–260 mm, locally exceeding 300 mm |
### 2. 2016-07-03 18:00 — Orange Rainstorm Warning (Continued)
| Item | Content |
|------|------|
| **Issue Time** | 2016-07-03 18:00 |
| **Warning Level** | Orange Rainstorm (Level II) |
### 3. 2016-07-04 06:00 — Orange Rainstorm Warning (Continued) ★Core
| Item | Content |
|------|------|
| **Issue Time** | 2016-07-04 06:00 |
| **Issuing Agency** | China Meteorological Administration (CMA) |
```

| **Warning Level** | Orange Rainstorm (Level II) |
| **Covered Areas** | **Central-southern Anhui**, **central-southern Jiangsu**, eastern Hubei, northern Hunan, northern Jiangxi, southeastern Guizhou, northern Guangxi |
| **Rainfall Forecast** | Localized heavy rainstorm 100–200 mm |
| **Remarks** | Orange Warning entering its 5th day |

### 4. 2016-07-04 — Major Meteorological Disaster (Rainstorm) Level II Emergency Response ★

| Item | Content |
|------|------|
| **Activation Time** | 2016-07-04 |
| **Activating Agency** | China Meteorological Administration |
| **Response Level** | **Major Meteorological Disaster (Rainstorm) Level II Emergency Response** |
| **Covered Areas** | Middle and lower Yangtze River region (Hubei, **Anhui**, **Jiangsu**, Guizhou) |
| **Trigger Condition** | Persistent heavy rainfall in the middle and lower Yangtze since June 30, with severe flooding in multiple areas |
| **Remarks** | Highest-level response of this round, second only to Level I |

### 5. 2016-07-04 18:00 — Orange Rainstorm Warning (Continued)

| Item | Content |
|------|------|
| **Issue Time** | 2016-07-04 18:00 |
| **Issuing Agency** | China Meteorological Administration (CMA) |
| **Warning Level** | Orange Rainstorm (Level II) |
| **Covered Areas** | **Central Anhui**, **central Jiangsu**, southeastern Hubei, northwestern Jiangxi, north-central Hunan, southeastern Guizhou, northern and southern coastal Guangxi |
| **Rainfall Forecast** | Heavy rainstorm in some areas |
| **Remarks** | From the 5th, the main rain belt will shift somewhat northward |

---

## Warning Evolution Timeline

```
2016-07-03 06:00   Orange Rainstorm (Continued)                    ← ★Within window
2016-07-03 18:00   Orange Rainstorm (Continued)                    ← ★Within window
2016-07-04 06:00   Orange Rainstorm (Continued, Day 5)             ← ★Within window ★Core
2016-07-04         Major Meteorological Disaster Level II Emergency Response   ← ★Within window ★Highest Response
                   (China Meteorological Administration)
2016-07-04 18:00   Orange Rainstorm (Continued)                    ← ★Within window
```

---

## Process Summary

This event marks the peak of the June 30 – July 6, 2016 heavy rainfall process in the middle and lower Yangtze River:

| Indicator | Value |
|------|------|
| **Orange Warning Duration** | 5 days (June 30 – July 4) |
| **Highest Warning Level** | Orange Rainstorm (Level II) |
| **Highest Emergency Response** | Major Meteorological Disaster Level II |
| **Maximum Cumulative Rainfall** | Locally exceeding 400 mm in eastern Hubei and central Anhui |
| **Heavy Rainstorm Coverage Area** | Over 140,000 $km^2$ across Hubei, Anhui, and Jiangsu |

| **Rainfall Period** | June 30 – July 6, 2016 |
---
## Related Events
Part of the concluding phase of the June 30 – July 6, 2016 historically severe rainfall process in the middle and lower Yangtze River:
- Event_06 (2016-06-30): First day of Orange Warning
- Event_07 (2016-07-01): Orange Day 2
- Event_08 (2016-07-02): Orange Day 3
- Event_09 (2016-07-03): Orange Day 4
---
*Data compiled: 2026-06-17 | Primary sources: CMA official website (cma.gov.cn), Xinhua News, China Meteorological Administration*

## S7. Reasoning outputs of AgenticAI-Control

The case reasoning outputs of AgenticAI-Control with various LLMs are presented as follows. All the other data are open-sourced on Github.

Github link: https://github.com/DantEzio/LLMAgent_UDSControl

Case of DeepSeek-V4-Flash:

Event ID: Event1
Time point: 2015-06-01 18:00
Estimated rainfall: Blue (Warning Continued).
Reasoning: The current simulation time corresponds to the warning issuance moment. A certain interval remains before rainfall onset, the water level at network nodes is extremely low, and the storage tank retains ample spare capacity. Following the principle of progressive intervention, Level 1 mild pre-emptive emptying is prioritized: only pump CC-S1 is activated to vacate tank volume in advance and avoid excessive energy consumption. Pump R remains shut down, as the condition of zero storage water level is not satisfied. All control actions comply with operational constraints (only binary state transitions from 0 to 1 are permitted).

Case of GLM-4-Flash:

Event ID: Event9
Time point: 2016-07-02 10:00
Estimated rainfall: Orange (Warning Continued).
Reasoning: As the orange rainstorm warning remains in force with substantial total rainfall and high peak rainfall intensity anticipated, the emptying pump shall be activated in advance to reserve storage tank capacity in accordance with BC control rules and the principle of progressive intervention, while Pump R remains unchanged.

Case of Moonshot-v1-8k:

Event ID: Event10
Time point: 2016-07-03 18:00
Estimated rainfall: Orange (Warning Continued).
Reasoning: According to the rainfall early warning information, the predicted total rainfall depth reaches 30.0 mm with a peak rainfall intensity of 13.0 mm/hr and a total duration of 6 hours. The current simulation time is 18:00 on 3 July 2016, which coincides with the warning issuance timestamp (18:00:00, 3 July 2016), and the predicted rainfall onset is more than one hour away. In accordance with the BC control rules and pre-emptive emptying strategy, the emptying pump should be activated for pre-draining to reserve available storage capacity of the detention tank. Given the long lead time before rainfall arrival, Level 1 mild pre-emptive emptying is selected, with only pump CC-S1 turned on.

Case of MiniMax-M3:

Event ID: Event5

Time point: 2015-08-19 06:00

Estimated rainfall: Blue (Warning Continued).

Reasoning: The current simulation time is 06:00 on 19 August 2015. The rainfall peak is projected to occur at 23:00 (approximately 17 hours later), and rainfall onset is estimated at around 17:00, which is roughly 11 hours from the present moment—far exceeding the 1-hour pre-draining window. Although rainfall is still a long way off, a blue rainstorm warning has been issued for this event (30 mm within 6 hours), corresponding to moderate rainfall intensity. Water levels across all network nodes remain generally low, with a maximum value of only 0.156 m, and no urban flooding risk is identified at present. In line with the rule stating that intensive proactive pre-draining shall not be activated if rainfall is more than one hour away, it is inappropriate to aggressively turn on all S pumps. Nevertheless, considering the supplementary principle that pre-drainage pumps may run moderately under normal conditions together with the priority on flood safety, the Level_1 mild pre-emptive draining strategy is adopted. Only CC-S1 (the main pre-drainage pump for Chengguan District) is activated to vacate partial detention tank capacity in advance. This measure avoids excessive energy consumption and hydraulic pressure fluctuations in the pipeline network caused by premature over-draining. Pump R maintains its original operating state without adjustment.

## S8. Control performance and action of BC, AgenticAI-Control, and MPC in rainfalls Event4-10.

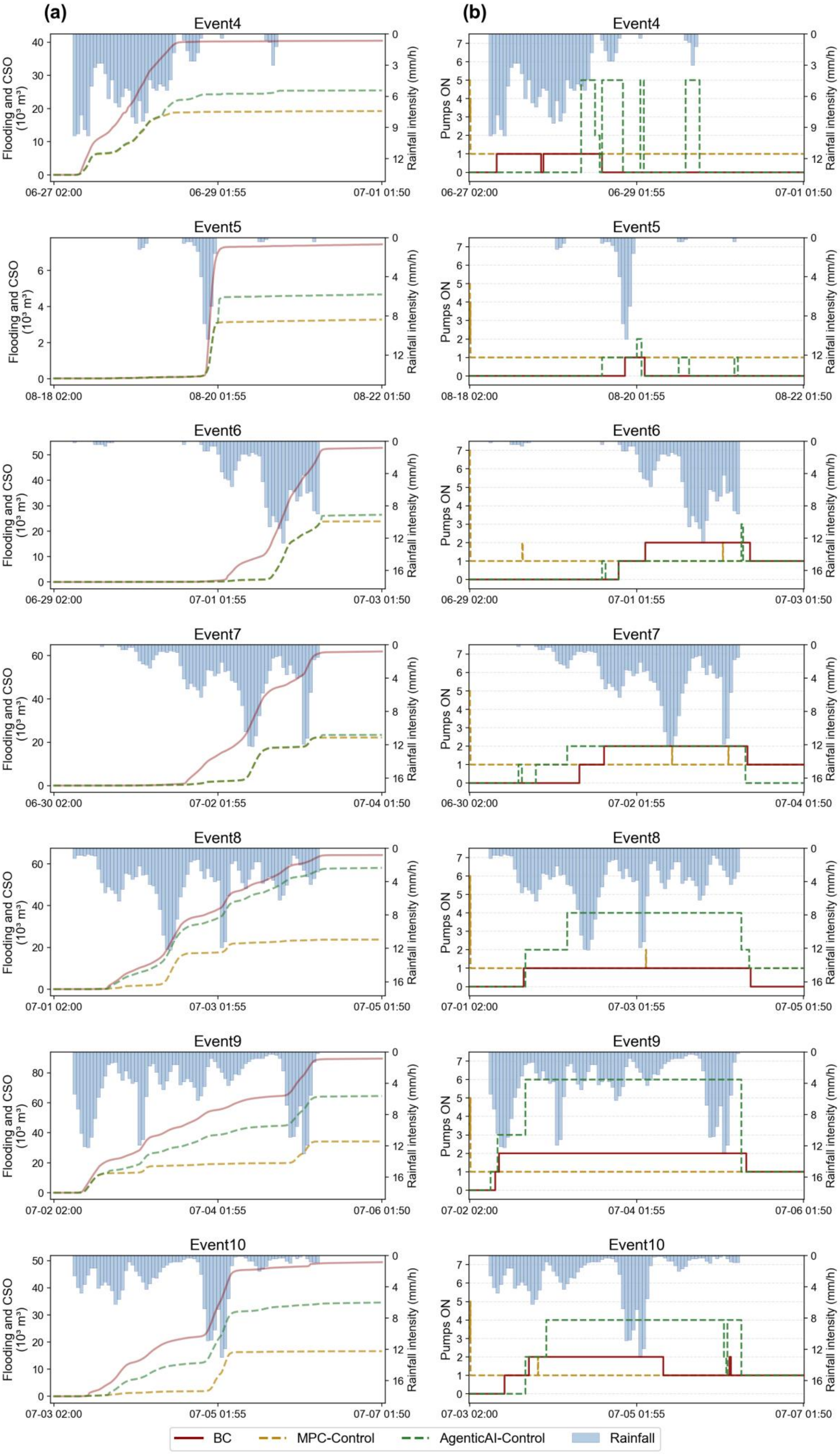

Fig. S3. Control performance and actions in Event4-10.

## S9. Control performance of AgenticAI-Control with four LLMs

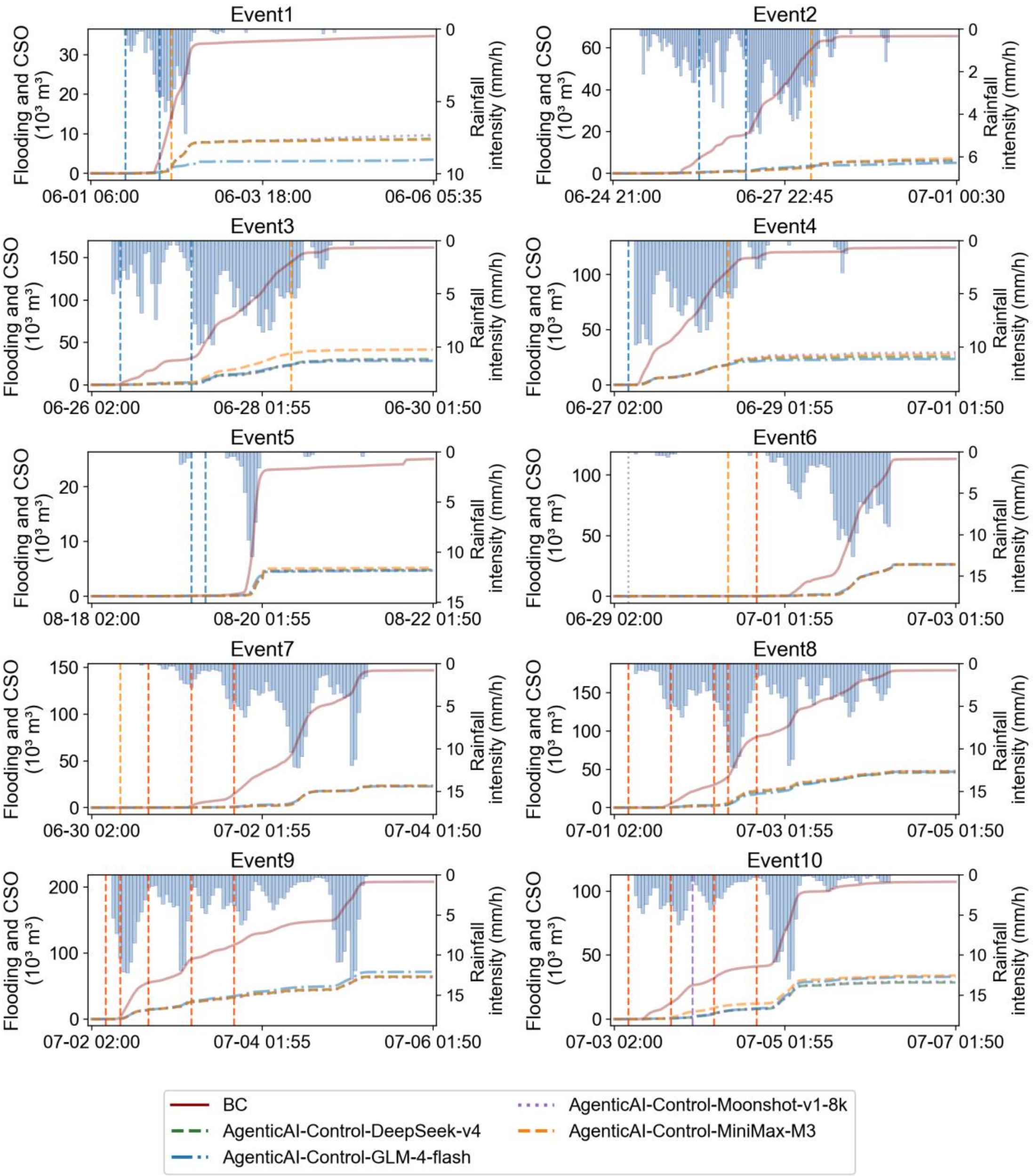

Fig. S4. Control process of AgenticAI-Control with various LLMs over 10 rainfall events.

## S10. Substitution term list

The synonyms used for substitution in uncertainty analysis are listed in Table S2. Since the study case is situated within China, both the original source texts collected and their alternative synonyms are in Chinese. Their English translations are also presented in Table S2.

Table S2. Substitution terms. Corresponding English translations are also provided.

| Category | Original Term | Synonyms for Replacement |
|---|---|---|
| Rainfall Intensity | 暴雨 heavy rain | [大雨，强降雨] [heavy rainfall, intense rainfall] |
| | 大暴雨 severe heavy rain | [特大暴雨，极强降雨] [extreme heavy rain, extremely intense rainfall] |
| | 雨强 rainfall intensity | [降雨强度，雨势强度] [rainfall magnitude, rain intensity] |
| | 雨带 rain belt | [降雨带，降水带] [rainfall belt, precipitation belt] |
| | 雨势 rain intensity trend | [降雨趋势，降水态势] [rainfall tendency, precipitation situation] |
| Early Warning Related | 预警 early warning | [警报，预警信号] [alert, warning signal] |
| | 降雨 rainfall | [降水，雨水] [precipitation, rainwater] |
| | 继续发布 continue issuing | [维持发布，持续发布] [maintain issuance, keep issuing] |
| | 升级 upgrade | [提升，调升] [elevate, raise] |
| | 发布 issue | [公布，通告] [announce, notify] |
| | 发布机构 issuing body | [发布单位，通告机构] [issuing entity, notifying agency] |
| Region / Coverage Scope | 覆盖区域 covered area | [影响范围，覆盖范围] [affected area, coverage area] |
| | 局地 localized area | [局部地区，部分地区] [partial area, some regions] |
| Prediction / Forecasting | 降雨预测 rainfall forecast | [降水预报，降雨预估] [precipitation forecast, rainfall estimation] |
| | 预报雨量 forecast rainfall volume | [预估雨量，预测降水量] [estimated rainfall volume, predicted precipitation] |
| Status Description | 维持 maintain | [保持，延续] [keep, continue] |
| | 增强 intensify | [加强，加剧] reinforce, exacerbate |

| | |
|---|---|
| 维持在 remain at | [保持在, 持续在] [stay at, persist at] |

**S11. MPC setting**

The MPC adopts a control interval of 10 minutes, and the prediction horizon for the optimization algorithm at each time step covers three timesteps, corresponding to a total duration of 30 minutes. Particle Swarm Optimization (PSO) algorithm is proposed to solve the optimization problem at every timestep, and the detailed parameter settings of PSO are listed in Table S3. All calculations in this work were performed solely on a 12th Gen Intel(R) Core(TM) i5-12600KF CPU with CPU-based parallelism enabled. Meanwhile, parallel computing was utilised to reduce the computing time of each optimization run. Since the computation time of the optimization for single control interval takes more than 10 minutes (Fig. 2 (b)), this MPC can only deliver optimal outcomes under ideal conditions

Table S3. The setting of PSO.

| Parameters | Value |
|---|---|
| Particle population | 10 |
| Iteration | 300 |
| Inertia weight | 0.8 |
| Acceleration coefficient | 0.5 |